\documentclass[opre,nonblindrev]{workingpaperJJB} % informs3

\OneAndAHalfSpacedXI

\usepackage{tracefnt}
\usepackage{anyfontsize}
\usepackage{amsmath}
\usepackage{ifthen}
\usepackage{enumerate}
\usepackage{url}
\usepackage{graphicx}
\usepackage[caption=false, font=tiny]{subfig}
\usepackage{setspace}
\usepackage{color}
\usepackage{booktabs}
\usepackage{makecell}
\usepackage{mathtools}
\usepackage{endnotes}
\let\footnote=\endnote
\let\enotesize=\normalsize
\def\notesname{Endnotes}%
\def\makeenmark{$^{\theenmark}$}
\def\enoteformat{\rightskip0pt\leftskip0pt\parindent=1.75em
  \leavevmode\llap{\theenmark.\enskip}}
\usepackage{float}
\usepackage[breaklinks]{hyperref}
\usepackage{url}
\usepackage[ruled,vlined,algo2e]{algorithm2e}
\usepackage{algpseudocode}
\usepackage{multirow}

\usepackage{tikz}
\usepackage{makecell}
\usepackage{hyperref} % See note above, other options in documentclass may be needed
\usepackage[T1]{fontenc}
\usepackage[utf8]{inputenc}
\usepackage{lmodern}
\usepackage{cfr-lm}
\usepackage{bbm}
\usepackage{array}
\usepackage{theorem}
\usepackage{cancel}
\usepackage{amsmath}
\usepackage{amsfonts}
\usepackage{amssymb}
\usepackage{graphicx}
\usepackage{longtable}

\newcommand{\E}{\mathbb{E}}
\renewcommand{\P}{\mathbb{P}}

\newcolumntype{C}{>{$}c<{$}}
\newcolumntype{L}{>{$}l<{$}}
\newcolumntype{R}{>{$}r<{$}}

\newcommand{\bg}{ \mathbf{g} }

\newcommand{\br}{ \mathbf{r} }

\newcommand{\bx}{ \mathbf{x} }

\newcommand{\by}{ \mathbf{y} }

\newcommand{\bmu}{ \boldsymbol{\mu} }

\newcommand{\btheta}{ \boldsymbol{\theta} }
\usepackage{soul}
\usepackage{natbib}
 \bibpunct[, ]{(}{)}{,}{a}{}{,}%
 \def\bibfont{\small}%

\newcommand\myineq[2]{\stackrel{\mathclap{\scriptstyle\mbox{#2}}}{#1}}

\TheoremsNumberedThrough 

\ECRepeatTheorems

\EquationsNumberedThrough

\RUNAUTHOR{Authors' names blinded for peer review}

\RUNTITLE{Algorithms for Nonstationary Combinatorial Bandits}

\TITLE{Finite-Time Guarantees for Multi-Agent Combinatorial Bandits with Nonstationary Rewards}

\ARTICLEAUTHORS{%
\AUTHOR{Katherine B. Adams}
\AFF{Department of Management Science and Statistics, University of Texas at San Antonio, San Antonio, TX 78249 \EMAIL{katherine.adams@utsa.edu}}
\AUTHOR{Justin J. Boutilier}
\AFF{Telfer School of Management, University of Ottawa, Ottawa, Canada, K1H 7V7 \\ \EMAIL{boutilier@telfer.uottawa.ca}}
\AUTHOR{Qinyang He, Yonatan Mintz}
\AFF{Department of Industrial and Systems Engineering, University of Wisconsin-Madison, Madison, WI 53706\\ \EMAIL{\{qhe57@wisc.edu, ymintz@wisc.edu\}}}
} % end of the block

\ABSTRACT{ 

We study a sequential resource allocation problem where a decision maker selects subsets of agents at each period to maximize overall outcomes without prior knowledge of individual-level effects. Our framework applies to settings such as community health interventions, targeted digital advertising, and workforce retention programs, where intervention effects evolve dynamically. Agents may exhibit habituation (diminished response from frequent selection) or recovery (enhanced response from infrequent selection). The technical challenge centers on nonstationary reward distributions that lead to changing intervention effects over time. The problem requires balancing two key competing objectives: heterogeneous individual rewards and the exploration-exploitation tradeoff in terms of learning for improved future decisions as opposed to maximizing immediate outcomes. Our contribution introduces the first framework incorporating this form of nonstationary rewards in the combinatorial multi-armed bandit literature. We develop algorithms with theoretical guarantees on dynamic regret and demonstrate practical efficacy through a diabetes intervention case study. Our personalized community intervention algorithm achieved up to three times as much improvement in program enrollment compared to baseline approaches, validating the framework's potential for real-world applications. This work bridges theoretical advances in adaptive learning with practical challenges in population-level behavioral change interventions.
}


\begin{document}

\maketitle

% - However, recent advances in data availability allow for new opportunities to personalize and improve the efficiency/effectiveness of thse services
    % - Need lots of data at the individual level to do this
% - We develop methods that can take advantage of this data for situations with X/y characteristics
% - Inspiration for this work

% Settings where influencing human behavior is crucial to operational success. 

% how operational decisions must take into account the impact on heterogeneous individuals 
% uniqueness about each decision-making individual

% tease out

% We are allocating a limited resource (only so many each period) to individuals over time, we don't know the impact of providing or not providing the resource but we learn over time,
% dynamically adapt our decisions as new information/feedback is obtained 

%%%%%% INTRODUCTION %%%%%%
\section{Introduction}

% Paragraph 1: Brief introduction to problem
Optimizing systems where outcomes are driven by human behavior is a critical challenge in operations research that includes applications such as behavioral health interventions \citep{mintz2017behavioral,liexpress,adams2023planning,he2024non}, online advertising \citep{afeche2017customer, furman2021customer}, and workforce retention \citep{jaillet2022strategic, arlotto2014optimal}. In these settings, a central decision maker must tailor resource (e.g., financial incentives, one-on-one counseling, goal setting, etc.) allocations to heterogeneous individuals (agents) to maximize the desired operational, health, or financial population outcomes. There are two key challenges faced by a decision maker in this setting. First, the impact of a resource allocation on different individuals is not known to the decision makers with certainty \emph{a priori} and must be learned through repeated observations. This leads to what is known in the reinforcement learning (RL) and multi-armed bandit (MAB) literature as the exploration vs. exploitation tradeoff, where the decision maker must consider whether to use a resource allocation that can assist them in learning more about the system or whether to use an allocation that maximizes the population outcome given current best estimates. This is further exacerbated in this setting, where the decision may be to withhold a resource from an agent that would seemingly benefit from it based on current estimates, in order to learn how such a resource may impact the behavior of a different agent whom the decision maker is more uncertain about. The second challenge is that the impact of these resource allocations may change depending on previous decisions due to frequent shifts in individual states and environmental context. This form of nonstationarity is common in the settings we described above and must be accounted for. Thus, there is a critical need for modeling frameworks that can assist decision makers in navigating these challenges to effectively allocate scarce resources while learning the parameters of their heterogeneous agent cohorts. %leading to the need to learn the agents' behaviors over time. 
%While modeling dynamic individual behavior is challenging, increasing data availability allow for new opportunities to personalize and improve the effectiveness of these systems.

% Paragraph 2: More detailed problem characterization
To address this need, we focus on the setting of nonstationary combinatorial MAB problems. In our setting, the decision maker is tasked with allocating limited resources to maximize a global outcome that is a function of the outcomes of individual agents as described above. The main decision per agent is binary, either to allocate a resource or not to allocate one. Therefore, the decision maker must choose a subset of agents to allocate resources to without exceeding their total budget. A key to our setting is that the the underlying states driving agents' responses are stochastic with unknown distribution, can change frequently over time, and are affected by the prior allocation of resources. Hence, the set of agents to be selected to maximize the desired outcome may change at each period. Our goal is to devise policies to guide the decision maker when determining to whom resources must be allocated at each period to maximize their global objective. 

%sequential decision problems where a central decision-maker is tasked with allocating limited resources to maximize a desired global outcome that is a function of the outcomes of individual agents. The decision-maker can affect agent behavior via their actions, where an action refers to whether a resource is allocated to an agent at a given period. To make effective decisions, the decision-maker must learn the impact of their actions on the agents' behaviors over time and dynamically adapt to the feedback received. We assume that the underlying states driving agents' decisions can change frequently over time and are affected by the provider decisions to allocate or withhold resources, with each of these two actions leading to distinct agent trajectories. Hence, the set of agents to be selected to maximize the desired outcome may change at each period. Our goal is to devise policies to guide the central decision-maker when determining to whom resources must be allocated at each period. 

\subsection{Application to Community Health Worker Interventions}
Our work is motivated by the real world problem of planning community health worker (CHW) interventions for diabetes treatment. In 2021, approximately 536.6 million people from 20 to 79 years old (10.5\%) were affected by diabetes globally, a number that is expected to increase to 783.2 million (12.2\%) in 2045 \citep{sun2022idf}. In countries and regions affected by a shortage of healthcare workers, employing CHWs provides an effective, affordable, and culturally-tailored approach to expand health system capacity for diabetes care through task shifting \citep{world2018guideline}. However, building effective treatment plans for CHW interventions is a challenging problem and highlights three characteristics that motivated our work. 

First, the decision-maker must consider how the timing of resource allocation impacts outcomes. In the case of CHW interventions, we need to learn how the frequency of CHW visits impacts each patient's propensity to engage in treatment. For example, if individuals are seen too often, they may become overwhelmed with information and disengage from treatment; on the other hand, if they are not seen often enough, they may not experience the benefits from treatment and also disengage. Second, the decision-maker must consider the constrained combinatorial nature of the problem because only a subset of individuals can receive resources at each period. The number of visits in each period is limited by the number of CHWs, therefore decision makers must use methods that allow them to effectively prioritize which participants to visit. Algorithmically, this also means that there is an exponential number of possible assignments of CHWs that must be sifted through in a tractable manner. Finally, we must manage the exploration-exploitation tradeoff. To do so, we must determine whether to allocate resources to improve outcomes now or to learn more about the system (potentially at a current cost), which may allow us to make better decisions in the future. For example, in practice, we are unable to observe an individual's health state (e.g., blood glucose) unless we allocate a CHW to visit them so we may want to visit individuals solely to obtain information on their current health state.   

%%
%%%%WHAT ARE OTHER EXAMPLE PROBLEMS BEYOND CHWS??
%%%
\subsection{Contributions}
In this paper, we develop a framework for combinatorial nonstationary MAB problems. %we develop a framework for allocating CHW appointments for a behavioral diabetes intervention delivered to workers in office settings within LMICs. 
Our framework provides three major contributions:
\begin{enumerate}
    %\item We extend existing models in the literature to incorporate features relevant to the implementation of CHW-delivered personalized wellness programs for chronic diseases under partial information (Section~\ref{sec:model-framework}). Our model accounts for disease progression and worker engagement with the program based on their health and motivational states. Additionally, it trades off screening and follow-up appointment decisions, as well as care provision based on current knowledge and information gathering to improve future decisions.
    \item We provide novel algorithmic approaches to solve nonstationary combinatorial multiarmed bandit problems. In particular, we develop Upper Confidence Bound-style algorithms to maximize outcomes (Section~\ref{sec:policies-cnbp}). We present theoretical analysis showing that our policies achieve regret of the order $\mathcal{O}\big(\log (T)\big)$ under mild assumptions. This a significant result because model-free approaches have only been shown to achieve square root regret and $\mathcal{O}\big(\log (T)\big)$ is the best bound achieved for nonstationary and for combinatorial bandits (separately).
    \item We provide a tuned version of our algorithm for future implementation in empirical simulation experiments (Section~\ref{sec:experiments}). Although it does not offer the same theoretical guarantees, our tuned algorithm uses asymptotic bounds that have been previously shown to have faster convergence in practice. We evaluate this approach in computational experiments against three existing methods from the literature.  Our results show that not only is our approach able to achieve sublinear cumulative regret (while existing approaches do not), but it is also able to achieve 20\% more average reward than the best competing baseline approach.%DETAILS ON ACTUAL PERFORMANCE?
    \item We conduct a comprehensive case study using real data from a CHW program for diabetes in India. In this setting, we extend existing models in the literature to incorporate features relevant to the implementation of CHW programs for chronic diseases with partial information (Section~\ref{sec:case-study}). We use this data to conduct simulation experiments that validate our model by comparing it to existing approaches in the literature (Section~\ref{sec:experiments}). Our experiments show that our approach achieves three times greater average program enrollment compared to the best baseline approach (Section~\ref{sec:exp-results}).
\end{enumerate}

\section{Literature review}
\label{sec:lit-review}

Our work contributes to three streams of literature: sequential decision-making for resource allocation (Section~\ref{sec:seq-res-alloc}), personalized healthcare (Section~\ref{sec:pers-health}), and global health operations (Section~\ref{sec:global-health}).

\subsection{Sequential decision-making for resource allocation under uncertainty}
\label{sec:seq-res-alloc}

%POMDPS; MABs (combinatorial and nonstationary)

Two main frameworks are employed when studying Sequential Decision-Making (SDM) problems under uncertainty: Markov Decision Processes (MDPs) and variants of the Multi-Armed Bandit (MAB) problem. %These two frameworks have been applied to a wide-range of settings, such as machine and infrastructure maintenance \citep{ruiz2020multi, van2020infrastructure}, online retail \citep{gao2022joint, boutilier2016budget}, product development \citep{bhaskaran2021sequential}, inventory management \citep{zhang2020closing}, and smart grid control for EV charging \citep{de2017complexity}. Within healthcare, SDM models have been built and used for noncommunicable diseases \citep{denton2018optimization}, including cancer screening \citep{lee2019optimal, ayvaci2012effect, ayer2012or}, liver transplantation \citep{alagoz2004optimal}, glaucoma monitoring \citep{helm2015dynamic}, stress and depression treatment \citep{aung2022planning}, among others. SDM models have also been used for infectious diseases, including HIV treatment \citep{shechter2008optimal}, COVID-19 vaccination \citep{rao2022sequential}, hepatitis C treatment \citep{deo2022optimal}, tuberculosis treatment \citep{mate2020collapsing}, and sepsis prediction \citep{liu2022machine}.
%MDPs are studied in control theory, where decision makers model systems that evolve stochastically over time to maximize the cumulative (often discounted) reward. The controller can affect the system via their action selection since the current state and chosen action both affect transition probabilities to future states. The controller's goal is to obtain optimal policies that map known states to optimal actions \citep{puterman2014markov, Poupart2010}. Partially Observable MDPs (POMDPs) are an extension of MDPs that were first introduced to the field of Operations Research in the 1960s \citep{drake1962observation, astrom1965optimal}. As opposed to MDPs, where actual system states are known, optimal policies for POMDPs map belief states to actions.
Within the MDP literature, our problem is closest to the study of constrained multiagent POMDPs \citep{de2021constrained}, weakly-coupled POMDPs \citep{parizi2019weakly}, and weakly-coupled multiagent MDPs \citep{gagrani2020weakly}. These problems are characterized by settings where a central decision maker must plan for shared resource usage of individual components or agents under soft or hard budget constraints. As noted by \cite{de2021constrained} in their survey, algorithms that can handle partial observability and hard budget constraints simultaneously are lacking in the operations literature. Both of these modeling features are essential for our problem, therefore we require the development of novel algorithmic approaches that are scalable and provide performance guarantees. %We next consider the literature on MABs.

The MAB problem is part of Reinforcement Learning (RL), a general framework where a decision maker attempts to optimize decisions by learning parameters of an underlying stochastic process \citep{sutton2018reinforcement}. %The classical version of the MAB problem consists of several rounds of play where an agent seeks to maximize their total reward by selecting a single arm from a set of arms that have stationary, stochastic, and time-independent reward distributions with unknown means \citep{berry1985bandit}. At each round, the agent receives some form of feedback (reward) from the chosen arm. The goal of a MAB algorithm is to decide which arm to play next based on the outcomes of the previous rounds in order to minimize its \emph{regret} -- the difference in total rewards from a clairvoyant (optimal) policy and the policy given by the algorithm. The MAB problem models a fundamental tradeoff in artificial intelligence between exploration and exploitation, where an agent has to balance reaping rewards from arms that have been previously identified as good options, and acquiring information on the rewards distributions of other arms that may yield even better rewards. Analytical tools for online learning such as those used in bandit problems are foundational for finite-sample analysis (i.e., convergence rate) in RL.
Many variants of the classical MAB problem have been introduced to better fit real-world settings, e.g., versions that incorporate stochastic, adversarial, or Markovian payoff evolution; linear, Lipschitz, or Gaussian payoff functions; full, bandit, or semi-bandit feedback; and side/contextual information -- to name a few (see \cite{bubeck2012regret} for an overview). Furthermore, some variants allow the selection of several arms per round \citep{chen2013combinatorial, gai2012combinatorial, das2022budgeted}, the number of arms that a player must choose from to be infinitely many \citep{wang2008algorithms, rusmevichientong2010linearly}, and for reward distributions to shift over time \citep{he2024non, mintz2020nonstationary, zhao2020simple}. For a review of recent advances in MAB problems for SDM, we refer the reader to \cite{agrawal2019recent}. In contrast with existing frameworks, our problem setting requires the use of bandits with shifting reward distributions (nonstationary bandits), side information (contextual bandits), and that allow the selection of multiple arms per round (combinatorial bandits). We provide additional information on MAB variants pertinent to our work and theoretical analyses in Section~\ref{sec:background-bandits}. 

% \ka{Cite \citep{mintz2018control} somewhere}

\subsection{Personalized healthcare}
\label{sec:pers-health}

%Personalized screening and treatment

%Patient agency

%The emergence of predictive models for health care have allowed providers to improve patient health outcomes by estimating health risks and treatment effects at the individual-level. These predictive models are useful for screening, treating, and monitoring patients for/with several health conditions. Prior work in personalized medicine has applied extensions of MDPs to strategically screen patients by accounting for pre-existing health conditions \citep{hajjar2022personalized}, treating diseases that have stochastic lead times for treatment initiation (such as chronic kidney disease) \citep{skandari2021patient}, monitoring cardiac events under energy constraints using mHealth technologies \citep{yao2021constrained}, treating hypertension with intuitive and scalable Poisson approximations that closely match optimal policies \citep{schell2016data}, predicting postpartum blood pressure spikes \citep{infusino2024predictive}, and treating hypertension while accounting for medication dosages and side effects \citep{choi2017dynamic}. 

MAB models for personalized healthcare have primarily focused on treatment \citep{keyvanshokooh2019contextual,negoescu2018dynamic,wang2019tumor,bastani2020online}. Some recent work has focused on facilitating patient engagement in treatment by modeling them as decision-making agents \citep{mintz2020nonstationary,rojas2021exploring,zhou2023spoiled}. In contrast, our problem setting requires modeling both provider and patient decisions since program participants have inherent motivational states that affect the intervention effectiveness. In this regard, our work is similar to that of \cite{liao2020personalized}, who developed a RL algorithm to improve just-in-time adaptive interventions to increase physical activity among individuals with hypertension, and \cite{aswani2019behavioral}, who developed behavioral models to improve participant adherence to weight loss interventions by characterizing patients' responses to changes in the system's states and inputs. \cite{mintz2017behavioral} provide an extension of the latter for behavioral interventions where a centralized decision-maker chooses financial incentives for intervention participants to maximize the treatment effect. 

%ROGUE/mintz2017
%Lacking the combinatorial (on/off) component that we have, which requires "choosing" who to visit, not the amount of incentive to allocate
%Leads to a trade-offs between who to visit / who not to visit (several implications 

%\cite{adams2023planning} 
%Partial information extension, which means that if we don't visit a person we don't know anything about them. Prior paper, we assumed we roughly know this (clustering, etc.)

%Our problem has an additional attribute that these frameworks cannot accommodate, that is, the provision of two types of visits (screening and management). Due to the lack of information for employees that are yet to be screened, we require a model where exploration (screening new program participants) and exploitation (treating enrolled participants) decisions can be traded off explicitly (\jb{see Section X for details}).

%simultaneously consider contextual information, nonstationary reward distributions, and budgeted decisions.

\subsection{Global health operations}
\label{sec:global-health}

%Operational problems in resource-limited settings

%CHW visit planning

Our work also contributes to the the growing literature stream of global health operations, a field concerned with optimizing care provision in resource-limited settings to achieve health equity worldwide \citep[e.g.,][]{kao2023strengthening, de2022vaccine, jonasson2017improving,parvin2018distribution,boutilier2020ambulance,boutilier2022improving}. For a review of operations research in global health, we refer the reader to \cite{bradley2017operations}. Past global health research on improving access to care has focused primarily on location and routing decisions for community-based or mobile clinic care \citep{brunskill2010routing,cherkesly2019community,oliphant2022improving,de2021site,santa2023multi}. In relation to these articles, our problem does not need to account for routing decisions since program participants are concentrated in the workplace (see Section~\ref{sec:case-study} for details). 
In this regard, our problem is similar to the work of \cite{adams2023planning}, who developed a SDM framework to plan CHW visits for patients living with diabetes in densely populated areas in LMICs (e.g., urban slums). However, in contrast to their framework, we require a RL-style framework that is capable of treating participants while learning more about them to improve decisions over time (i.e., achieve sublinear regret).

%%%%%% MODEL DESCRIPTION %%%%%%
\section{Model description}
\label{sec:model-framework}

In this section, we present a novel framework to make sequential decisions in resource-limited settings, which we denote by Combinatorial Bandits with Recovery and Habituation (COBRAH). We apply this framework to problems with nonstationary reward distributions and binary actions for each component of the system. An action for a single component (arm selection) corresponds to the decision to allocate a resource to that component. Due to limited resources, we assume that only a subset of arms may be selected at each period. In Section~\ref{sec:background-bandits}, we provide background information on bandit problems that are relevant to understand our theoretical analyses, in Section~\ref{sec:comb-ns-prob}, we present our general modeling framework, in Section~\ref{sec:technical-assumptions} we describe technical assumptions, and, in Section~\ref{sec:traj-kl-div}, we briefly introduce a measure for comparing estimated and true reward distributions.
%Section~\ref{sec:case-study-framework}, we show our model for the particular application of a provider planning CHW visits to screen and treat patients with diabetes under partial information. 

% Based on the results from our previous paper, the goal of improving glycemic control can be achieved by maximizing the number of patients enrolled in treatment across the planning horizon.
%increase glycemic control among their employees. 

\subsection{Background on Stationary and Resource-Constrained MABs}
\label{sec:background-bandits}

% General stationary MAB problem, regret definition

% Noteworthy results and UCB intro

% KL divergence

In the stationary MAB problem, a decision-maker must choose an action $a$ from a finite set $\mathcal{A}$ at each period $t \in \mathcal{T}$ to maximize a stochastic reward $r_{a,t}$. Rewards are assumed to be i.i.d  with a fixed sub-Gaussian distribution $\mathbb{P}_{\theta_a}$, where parameter $\theta_a$ is unknown but assumed to lie in a known compact set $\Theta$ \citep{lai1985asymptotically}. The goal is to maximize the expected reward by constructing a policy $\pi=\{\pi_t(\cdot)\}_{t \in \mathcal{T}}$, that is, a sequence of decision mappings based on past rewards and actions. This problem can be written as $\max_{\pi \in \Pi} \sum_{t=1}^T\mathbb{E}[r_{\pi_t}]$, where $\Pi$ is the set of all admissible policies. 

A common way to evaluate algorithmic performance in the MAB literature is using a measure called regret, which is the difference between the rewards of an optimal action and of the actions chosen by a  policy $\pi$ ($r_{a^*}-r_{\pi_t}$, where $a^* \in \argmax_{a \in \mathcal{A}} \E[r_a]$). The main metric of interest is cumulative expected regret  defined as $\E\big[R_\Pi(T)\big]=\E\Big[T \cdot r_{a^*}-\sum_{i=1}^T r_{\pi_t}\Big]$. Note that minimizing $\E\big[R_\Pi(T)\big]$ is equivalent to maximizing $\sum_{i=1}^T \E[r_{\pi_t}]$. In general this measure is able to capture how quickly a policy learns the true parameters of the system and is  able to trade off exploration and exploitation. Trivially, any policy will be able to achieve linear regret; however, effective policies will be able to achieve sublinear regret, indicating that they choose suboptimal actions less often as time elapses.
%
%To determine how much an algorithm has learned after $T$ rounds of play (periods) based on past actions and rewards, its performance is evaluated using the cumulative expected regret, defined as $\E\big[R_\Pi(T)\big]=\E\Big[T \cdot r_{a^*}-\sum_{i=1}^T r_{\pi_t}\Big]$. Note that minimizing $\E\big[R_\Pi(T)\big]$ is equivalent to maximizing $\sum_{i=1}^T \E[r_{\pi_t}]$. Using the expected regret allows for stochasticity not to be factored into the algorithm's performance and to measure how much the algorithm is learning from experience. Intuitively, it is desirable that algorithms reduce the regret per period (increase rewards) as more data is collected from past actions. In theoretical analyses, the goal is to show that the cumulative expected regret of a proposed algorithm is sublinear, which indicates that it is approaching the optimal decision (i.e., learning) as the number of plays increases. Regret proofs rely on two core steps: bounding the suboptimality of a bad play and the number of bad plays. 
%The number of bad plays is bounded via the use of concentration inequalities, which bound the probability of an extreme event, i.e., an event occurring on the tails of a distribution \citep{lattimore2020bandit}. 

Among pertinent theoretical results for the stationary MAB problem, we highlight \cite{lai1985asymptotically}, who showed that logarithmic regret can be achieved asymptotically; \cite{auer2002finite}, who showed that logarithmic regret can be achieved not only asymptotically, but uniformly over time; \cite{Gittins:89}, who showed that a policy based on dynamic allocation indices (known as the Gittins index) is optimal for the infinite-horizon discounted stationary MAB; and \cite{maillard2011finite}, who provided finite-time analyses for the stochastic MAB problem with finite-dimensional parametric distributions using a measure of statistical distance called the Kullback-Leibler (KL) divergence, paving the way for handling general finite-dimensional parametric distributions and incorporating empirical reward distributions. Among common algorithmic approaches, upper confidence bounds (UCBs) have been proposed to take actions using the principle of ``optimism in the face of uncertainty". For instance, the UCB1 algorithm has been shown to achieve logarithmic regret under the bounded rewards assumption for the stationary MAB problem \citep{auer2002finite}. 
% Cite Gittins here? It is relevant but may seem out of place

More recently, authors have explored stationary bandits that have a resource constraint. For instance, \cite{chen2013combinatorial} developed algorithms with $\mathcal{O}\big(\log (T)\big)$ regret for combinatorial bandits, where the number of possible actions grow exponentially with the number of arms. With algorithms defined under mild assumptions (monotonicity and bounded smoothness), their results can be broadly used, including in problem settings that involve nonlinear rewards. In the case of linear rewards, they significantly improve on the bound of \cite{gai2012combinatorial}. Additionally, \cite{badanidiyuru2018bandits} introduced the bandit with knapsacks (BwK) problem, where decisions taken at each time period of a finite horizon have stochastic resource consumption (subject to budget constraints) and reward generation. In their case, the optimal policy may not be to play the arms with greatest expected rewards at every period since the resource consumption of each action and the remaining supply of each resource must be considered. The authors define an optimal dynamic policy and two solution algorithms that achieve optimal regret up to polylogarithmic factors. \cite{sankararaman2017combinatorial} define a generalization for both BwK and combinatorial semi-bandits denoted by combinatorial Semi-Bandits with Knapsacks (SemiBwK), where rewards are observed only for chosen arms. Their algorithms assume the action set is described by a matroid constraint and achieve regret comparable to both BwK problems and combinatorial semi-bandits.

\subsection{Combinatorial Bandits with Recovery and Habituation (COBRAH)} \label{sec:comb-ns-prob}

In this subsection, we describe details of our COBRAH framework. Let $\mathcal{A}$ denote the set of finite arms that may be chosen at each period (e.g., the set of patients in the community or business targeted by the CHW intervention). Let $\mathcal{T}=\{1,\dots,n\}$ be the planning horizon. The state of arm $i \in \mathcal{A}$ at time $t \in \mathcal{T}$ is denoted by $x_{i,t} \in \mathcal{X}$, the arm parameters by $\theta_i \in \Theta$ (where $\mathcal{X},\Theta$ are known compact and convex sets), and the decision to select action $i$ at time $t$ by $y_{i,t} \in \{0,1\}$. In contrast to the classical MAB problem, we assume that multiple arms may be selected at each period. We assume state $x_{i,t}$ evolves according to dynamics of the form $x_{i,t+1}=f_i(\theta_i, x_{i,t}, y_{i,t})$, where $f: \mathcal{X}\times \mathbb{B} \xrightarrow{} \mathcal{X}$ is a known function. We use parameter $C$ to represent the maximum number of arms that can be chosen at each time period and write the capacity constraint as follows:
\begin{align}
    \sum_{i \in \mathcal{A}} y_{i,t} \leq C \quad \forall t \in \mathcal{T}
\end{align}
% which we can rewrite in terms of the super-action $Y_{i,t}$, where $\mathcal{S}$ denotes all possible subsets of patients (super-arms) that could be visited (pulled) in a period.
% \begin{align*}
%     \sum_{i \in \mathcal{P}_j} Y_{i,t} \leq C \quad \forall j \in \mathcal{S}, t \in \mathcal{T}
% \end{align*}

%Note: "Old" Value of enrollment and Enrollment rewards sections are commented out due to decision to use semi-bandit feedback case instead of full information feedback

%Individual arm states evolve according to dynamics of the form
% \begin{align}
%     x_{i,t+1} &= f(x_{i,t}, y_{i,t}, \xi_{i,t}, \zeta_{i,t})
% \end{align}
% where $\xi_{i,t}$ is the random disturbance for patient $i$ at time $t$. 
Therefore, at each period $t$, there are $2^C$ subsets of arms that we may select. We denote each possible subset of arms that may be selected as a super-arm $S \in \mathcal{S}$, where $\mathcal{S}$ is the set of all super-arms. Each arm has an unknown stochastic reward $r_{i,t}$  with mean $\mathbb{E}[r_{i,t}]  = g(\theta_i, x_{i,t})$, where $g(\cdot)$ is a bounded function. At some points, for brevity, we will use the shorthand $g_{i,t} = g(\theta_i,x_{i,t})$. We assume $x_{i,t}$ can be affected by action $y_{i,t}$ whether arm $i$ is selected in period $t$ or not (restlessness), and, therefore, $\by_t=(y_{1,t}, \dots, y_{|\mathcal{A}|, t})$ can indirectly influence the rewards for the same period, $\br_t = [r_1,r_2,...,r_m]^\top$. Due to the nonstationarity, we require a regret definition that accounts for a non-static optimal policy. We use the expected cumulative dynamic regret (henceforth referred to as regret) since it accounts for contexts where the optimal policy $\pi_t^*$ changes at each period $t$. Our precise regret definition is provided in Section~\ref{sec:policies-cnbp}.

\subsection{Technical Assumptions} \label{sec:technical-assumptions}

We require some technical assumptions to obtain regret bounds for our solution algorithms:

\begin{assumption} \label{ass:cond-independence}
    Rewards $r_{i,t}$ are conditionally independent given initial conditions $x_{i,0}$ and parameters $\theta_i$.
\end{assumption}

This assumption is analogous to the independence of rewards assumption in the stationary MAB problem. It states that, for any two periods $t$ and $t'$, we have that $r_{i,t}|\{x_{i,t}, \theta_i\}$ is independent of $r_{i,t'}|\{x_{i,t'}, \theta_i\}$, that is, rewards are conditionally independent on states and parameters. Assumption~\ref{ass:cond-independence} allows us to express the likelihood function decoupling the reward probability based on the current state and parameters from the probability of being in a state based on the initial state and parameters, where the latter probability is calculated by applying the dynamics function $f$ from period 0 to $t$ based on past actions.

\begin{assumption} \label{ass:log-concave}
    The reward distribution has a log-concave probability density function $p(r|\theta,x)$ for all $x \in \mathcal{X}$ and $\theta \in \Theta$.
\end{assumption}

Assumption~\ref{ass:log-concave} provides regularity for the reward distributions and is met by common distributions such as the Gaussian and Bernoulli distributions. Assumption~\ref{ass:log-concave} guarantees the maximum of $p(r|\theta,x)$ is attainable, which we require to calculate its maximum likelihood estimator (MLE).

Prior to stating the next assumption, we define a function $f(\cdot)$ to be $L$-Lipschitz continuous on a compact set $\mathcal{D}$ if $\big|f(x_1) - f(x_2)\big| \leq L \|x_1 - x_2\|_2$ for all points $x_1, x_2$ of its domain on set $\mathcal{D}$.

\begin{assumption} \label{ass:log-lipschitz-cont}
    The log-likelihood ratio $\ell(r; \theta', x', \theta, x) = \log \frac{p(r|\theta', x')}{p(r|\theta, x)}$ associated with the distribution family $\mathbb{P}_{\theta,x}$ is locally $L_f$-Lipschitz continuous with respect to $x$ and $\theta$ on the compact set $\mathcal{X} \times \Theta$. We also assume that the mean reward function $g(\theta, x)$ is locally $L_g$-Lipschitz continuous with respect to $x, \theta$ on the compact set $\mathcal{X} \times \Theta$.
\end{assumption}

Assumption~\ref{ass:log-lipschitz-cont} ensures that any two probability distributions will be similar if their parameters have similar values. Along with Assumption~\ref{ass:dynamic-lipschitz}, Assumption~\ref{ass:log-lipschitz-cont} is used to show that the Euclidean distance between two log-likelihood ratios with different starting conditions and at different periods is within a radius of the size of a Lipschitz constant times diam$(\mathcal{X})$, where diam$(\mathcal{X})=\max_{x \in \mathcal{X}}\|x\|_2$.

\begin{assumption} \label{ass:sub-gaussian}
    The reward distribution $\mathbb{P}_{\theta,x}$ for all $\theta \in \Theta$ and $x \in \mathcal{X}$ is sub-Gaussian with parameter $\sigma$, and either $p(r|\theta, x)$ has finite support or $\ell(r; \theta', x', \theta, x)$ is locally $L_p$-Lipschitz with respect to $r$.
\end{assumption}

Assumption~\ref{ass:sub-gaussian} ensures that sample averages are close to their means. It is used to show that the maximum distance between the average log-likelihood ratio and trajectory KL divergence is Lipschitz with respect to the action sequence.

\begin{assumption} \label{ass:dynamic-lipschitz}
    The dynamic transition function $f(\cdot)$ is $L_f$-Lipschitz continuous such that $L_f \leq 1$.
\end{assumption}

Assumption~\ref{ass:dynamic-lipschitz} implies that the dynamics are stable, therefore states cannot change too quickly. Assumptions~\ref{ass:log-lipschitz-cont}, \ref{ass:sub-gaussian}, and \ref{ass:dynamic-lipschitz} are required to bound the suboptimality of each bad round.

\subsection{Preliminaries and Concentration Inequality}
\label{sec:traj-kl-div}

Key to the analysis of MABs is the notion of finite-time concentration inequalities. These inequalities give us a notion of how quickly a stochastic process concentrates around its mean as additional samples are collected \citep{keener2010theoretical,wainwright2019high}. In stationary bandit analyses, common inequalities used are Hoeffding's Inequality and Bernstein's Inequality \citep{lai1985asymptotically,auer2002finite}. Since our process of interest is nonstationary, we will instead use the following concentration inequality from \cite{mintz2020nonstationary}:
\begin{theorem}
    (Corollary 1 from \cite{mintz2020nonstationary}) For $\alpha \in (0,1)$:
    \begin{flalign}
    %\mathbb{P}\Bigg(\frac{1}{T_{i,t-1}}D_{i,\pi_1^t}(\theta_i,x_{i,0} || \hat{\theta}_i,\hat{x}_{i,0}) &\geq u_2 \Bigg) \leq u_3\\
    \mathbb{P}\Bigg(\frac{1}{T_{i,t-1}}D_{i,\pi_1^t}(\theta_i,x_{i,0} || \hat{\theta}_i,\hat{x}_{i,0}) &\geq B(\alpha)\sqrt{\frac{\log(1/\alpha)}{T_{i,t-1}}} \Bigg) \leq \alpha, \label{eq:mintz-conc-ineq}
\end{flalign}
where $B(\alpha) = \frac{c_f(d_x,d_{\theta})}{\sqrt{\log(1/\alpha)}}+L_p \sigma \sqrt{2}$ and $c_f(d_x,d_{\theta}) = 8 L_f \textnormal{diam}(\mathcal{X})\sqrt{\pi} + 48\sqrt{2}\cdot 2^{\frac{1}{d_x + d_{\theta}}} L_f \textnormal{diam}(\mathcal{X}\times\Theta)\sqrt{\pi(d_x + d_{\theta})}$, and $d_x$ and $d_\theta$ represent the dimensionality of $\mathcal{X}$ and $\Theta$, respectively.
\end{theorem}

% Prior to presenting our technical assumptions, we introduce a statistical measure from information theory called Kullback-Leibler (KL) divergence (also known as relative entropy). It is used to measure the excess ``surprise" from using distribution $Q(x)$ as a model instead of $P(x)$ when the true distribution is $P(x)$. It is asymmetric and therefore not a metric. The KL divergence is a nonnegative quantity that has value zero if and only if $P$ and $Q$ are identical, leading it to be used for several applications. For continuous random variables, is defined as:
% \begin{align}
%     D_{KL}(P || Q) &= \int_{-\infty}^{\infty} p(x) \log \bigg(\frac{p(x)}{q(x)}\bigg)
% \end{align}
% To compare probability measures for multiple periods, we require a measurement applicable to a trajectory of states and rewards.

Here $D_{i,\pi_1^t}$ represents the trajectory KL divergence defined in \cite{mintz2020nonstationary}, a measure that compares two joint probability distributions of rewards with the same input sequence $\pi_{1}^t=\{\by_1, \by_{2}, \dots, \by_{t-1}, \by_{t}\}$, same state dynamics function $f_i$, different parameters $\theta_i,\theta_i' \in \Theta$, and different starting conditions $x_{i,0},x_{i,0}' \in \mathcal{X}$. Algebraically, we can write this as:
\begin{align}
    D_{i,\pi_1^t}\big(\theta_i,x_{i,0} || \theta_i', x_{i,0}'\big) &= \sum_{t \in \mathcal{T}} D_{KL}\big(\mathbb{P}_{\theta_i,x_{i,t}}||\mathbb{P}_{\theta_i', x_{i,t}'}\big) = \sum_{t \in \mathcal{T}} D_{KL}\big(\P_{\theta_i, f_i^t(x_{i,0})} || \P_{\theta'_i,f_i^t(x'_{i,0})}\big)\\
    &= \E_{\theta_i, x_{i,0}}\Bigg[\sum_{t \in \mathcal{T}} \log \frac{p\big(r_{i,t}|\theta_i, f_i^t(x_{i,0},\theta_i,y_i)\big)}{p\big(r_{i,t}|\theta'_i, f_i^t(x'_{i,0}, \theta'_i,y_i)\big)}\Bigg]
\end{align}
where $f_i^t$ represents the functional composition of the dynamics function $f_i$ with itself $t$ times based on the input sequence $\pi_1^t$, and $D_{KL}$ represents the ordinary Kullback-Leibler divergence \citep{keener2010theoretical}.
% We also present a concentration inequality developed by \cite{mintz2020nonstationary} for the average trajectory KL divergence:
% \begin{flalign}
%     %\mathbb{P}\Bigg(\frac{1}{T_{i,t-1}}D_{i,\pi_1^t}(\theta_i,x_{i,0} || \hat{\theta}_i,\hat{x}_{i,0}) &\geq u_2 \Bigg) \leq u_3\\
%     \mathbb{P}\Bigg(\frac{1}{T_{i,t-1}}D_{i,\pi_1^t}(\theta_i,x_{i,0} || \hat{\theta}_i,\hat{x}_{i,0}) &\geq B(\alpha)\sqrt{\frac{\log(1/\alpha)}{T_{i,t-1}}} \Bigg) \leq \alpha, \label{eq:mintz-conc-ineq}
% \end{flalign}
% where $B(\alpha) = \frac{c_f(d_x,d_{\theta})}{\sqrt{\log(1/\alpha)}}+L_p \sigma \sqrt{2}$ and $c_f(d_x,d_{\theta}) = 8 L_f \textnormal{diam}(\mathcal{X})\sqrt{\pi} + 48\sqrt{2}\cdot 2^{\frac{1}{d_x + d_{\theta}}} L_f \textnormal{diam}(\mathcal{X}\times\Theta)\sqrt{\pi(d_x + d_{\theta})}$, and $d_x$ and $d_\theta$ represent the dimensionality of $\mathcal{X}$ and $\Theta$, respectively. 
%
The average trajectory KL divergence and the concentration inequality above will be used in our algorithms and theoretical results presented in Section~\ref{sec:policies-cnbp}.

\section{Policies for Optimizing COBRAH}
\label{sec:policies-cnbp}

In this section, we describe our solution approaches for the COBRAH problem. In Sections~\ref{sec:policies-sb} and \ref{sec:policies-ff}, we present our algorithms for the semi-bandit (SB) case and full feedback (FF) case, respectively. We also provide theoretical results for finite-time convergence of our two algorithmic approaches. In this section we provide sketches of the main results; however, the detailed proofs can be found in Appendix~\ref{app:proofs}.

% \subsection{Case Study Parameter Estimation} \label{sec:param-estimation}

%, and $f_\zeta:\mathbb{R} \mapsto \mathbb{R}_+$ is the density function for $\zeta$. 

%Note that we require upper and lower bounds on $B_t$, which we denote $L$ and $U$, respectively.

% \ka{Old:}
% Specifically, if $z_t \sim \text{Bernoulli}\Big(\frac{1}{1+\exp{(-B_t)}}\Big)$ then we can derive the log-likelihood as:
% \begin{align}
%     \log p(z_t|B_t) &= z_t \log\bigg(\frac{1}{1+\exp(-B_t)}\bigg) + (1- z_t) \log \bigg(\frac{\exp(-B_t)}{1+\exp(-B_t)}\bigg) \\
%     & = z_t \log\bigg(\frac{1}{\exp(-B_t)}\bigg) + \log \bigg(\frac{\exp(-B_t)}{1+\exp(-B_t)}\bigg) \\
%     & = z_t B_t - \log( 1 + \exp(B_t)) \label{eq:log_like}
% \end{align}

% Observe that the second term in \eqref{eq:log_like} can be rewritten as $-\log(1+\exp(B_t)) = -\log(\exp(0)+\exp(B_t)$, which yields an expression of the form log-sum-exp. We can apply the following inequalities:
% \begin{equation}
%     \max_{i \in \{1,...,n\}} (x_i) \leq \log \sum_{i=1}^n \exp{(x_i)} \leq \max_{i \in \{1,...,n\}} (x_i) + \log (n)
% \end{equation}
% Here we have that $n=2$, $x_1 = 0$, and $x_2 = B_t$, so that term can be reformulated as:
% \begin{equation}
%     \max\{0,B_t \}\leq \log\big(1 + \exp(B_t)\big) \leq \ \max\{0,B_t\} + \log (2)
% \end{equation}
% We obtain the following bounds on the term $-z_t B_t + \log\big(1+\exp(B_t)\big)$ in Eq.~\eqref{eq:mle_obj_original}:
% \begin{equation}
%     -z_t B_t + \max\{0,B_t\} \leq - \log p(z_t|B_t) \leq -z_t B_t + \max\{0,B_t\} + \log (2)
% \end{equation}

\subsection{Semi-Bandit Feedback}
\label{sec:policies-sb}

We first consider the case where rewards are observed only for arms selected in a given period, denoted in the MAB literature as semi-bandit (SB) feedback. We begin by presenting the algorithm itself and then its associated regret bound and proof.

\subsubsection{COBRAH-SB Algorithm} \label{sec:kl-cucb-sb-alg}

We develop a novel algorithm which we call COBRAH-SB for the SB version of problem introduced in Section~\ref{sec:comb-ns-prob}. Let $T_i$ be the number of times arm $i$ has been played and $(\theta_i, x_{i,0})$ be the parameters and initial conditions of arm $i$, respectively.  % include disturbance term? 
Let $f_{\pi}^t(x_{i,0})$ denote the functional composition of $f$ with itself $t$ times based on policy $\pi_1^t$, where each vector $\by_t$ has binary entries and size $|\mathcal{A}|$. The true expected rewards are denoted by $g(\theta_i, x_{i,0})$, and its estimates, by $g(\hat{\theta}_i, \hat{x}_{i,0})$.

The pseudocode for the COBRAH-SB algorithm is provided in Algorithm~\ref{alg:kl-cucb-sb} and it yields a policy for a finite planning horizon. The algorithm begins by playing each arm once (initialization) and then plays the remaining $n-m$ rounds using the UCBs calculated with the trajectory KL divergence. At each round, the MLE estimates for the $\theta_i$ and $x_{i,0}$ of all $i \in \{1,\dots,m\}$ are updated and the super-arm $S$ with the greatest sum of UCBs of expected rewards ($\sum_{i \in S} g_{i,t}^{\textnormal{UCB}}$) is played. Note that the optimal super-arm is indexed by $t$ since it may change at each time period.

% KL-CUCB -> Semi-bandit
\begin{algorithm2e}[ht]
    \linespread{1.5}\selectfont
    % \DontPrintSemicolon
    % \KwData{$(\by, \bz, \bar{\bb}, S, B, G, R)$}
    % \KwResult{$w^*$, $x^*$ = ($\hat{\bb}^*$, $\bs^*$, $\btheta^*$, $p^*$, $\mu^*$, $\alpha^*$, $\theta_0^*$, $\lambda^*$, $s_0^*$, $\beta^*$, $\gamma^*$, $\rho^*$)}
    \Begin{
        % Initialization rounds
        \For{$i \in \{1,\dots,m\}$}{
            $t\xleftarrow{}i$
            
            Play each $i$ once by choosing an arbitrary super-arm $S$ such that $i \in S$

            Update the number of times each arm $i$ has been played ($T_i$) and the vector of reward observations $\br_i = (r_{i,0}, \dots, r_{i,t})$ % z_{i,T_i}
        }
        \While{$t \leq n$}{
            $t \xleftarrow{} t+1$
            
            \For{$i \in \{1,\dots,m\}$}{
                Compute:
                
                $\hat{\theta}_i, \hat{x}_{i,0} = \argmin \Bigg\{- \sum\limits_{t \in \mathcal{T}_i} \log p(r_{i,t} | \theta_i, x_{i,t}) : x_{i,t+1} = f(\theta_i, x_{i,t}, y_{i,t}) \: \forall t \in \{0,\dots,n-1\} \Bigg\}$

                $g^{\textnormal{UCB}}_{i,t} = \max\limits_{\theta_i, x_{i,0} \in \Theta \times \mathcal{X}} \Big\{g\big(\theta_i, f_i^t(x_{i,0})\big) : \frac{1}{n(\mathcal{T}_i)} D_{i,\pi_1^t} \big(\theta_i, x_{i,0} || \hat{\theta}_i, \hat{x}_{i,0}\big) \leq B(t^{-4})\sqrt{\frac{4\log(t)}{T_{i,t-1}}} \Big\}$
            }
            Play $\Tilde{S}_t \in \argmax\limits_{S \in \mathcal{S}} \sum\limits_{i \in S} g^{\textnormal{UCB}}_{i,t}$; i.e., set $y_{i,t}=1 \:\forall i \in \Tilde{S}_t$ and $y_{i,t}=0 \: i \notin \Tilde{S}_t$

            Update $T_i \: \forall i \in \Tilde{S}_t$ and $r_{i,t} \: \forall i \in \{1,\dots,m\}$
        }
    
    }
    \caption{COBRAH-SB algorithm. \label{alg:kl-cucb-sb}}
\end{algorithm2e}

\subsubsection{COBRAH-SB Regret}

Let $\Tilde{\pi}=\pi_{1}^{n}$ be the complete policy given by the COBRAH-SB algorithm, $\pi^*$ denote the optimal action sequence obtained by a dynamic oracle that has full information of agent states at each time point, and $S^*_t \in \mathcal{S}_t^*$ be a super-arm in the set of optimal super-arms in period $t$. Because only the rewards of chosen arms factor into the total provider reward at each period, the regret of the algorithm is given by:
\begin{align*}
    R_{\Tilde{\pi}}(n) = \sum\limits_{t=1}^n \Bigg(\sum\limits_{i \in S_t^*} g\big(\theta_i, f_{\pi^*}^t(x_{i,0})\big) - \sum\limits_{i \in \Tilde{S}_t} g\big(\theta_i, f_{\Tilde{\pi}}^t(x_{i,0})\big)\Bigg),
\end{align*}
where $\Tilde{S}_t$ is the super-arm selected by policy $\Tilde{\pi}$ in period $t$. Our key result provides an upper bound on the regret $R_{\Tilde{\pi}}(n)$ of the policy obtained with the COBRAH-SB algorithm:
\begin{theorem}[Regret bound] \label{thm:klcucb-regret}
    The expected regret after $n$ rounds of play for policy $\Tilde{\pi}$ obtained using the COBRAH-SB algorithm is bounded by
    \begin{align*}
        \E[R_{\Tilde{\pi}}(n)] \leq C L_g \textnormal{diam}(\mathcal{X} \times \Theta)  |\mathcal{S}|\Bigg(\frac{4\big(B(\lceil \frac{m}{C}\rceil^{-4})\big)^2\log(n)}{\delta_{\min}^2} + \frac{m^2\pi^2}{3}\Bigg),
    \end{align*}
    where  
    \begin{align}
        &B(\alpha)= \frac{c_f(d_x,d_{\theta})}{\sqrt{\log(1/\alpha)}}+L_p \sigma \sqrt{2},\\
        &c_f(d_x,d_{\theta}) = 8 L_f \textnormal{diam}(\mathcal{X})\sqrt{\pi}+48\sqrt{2}(2)^{\frac{1}{d_x + d_{\theta}}}L_f\textnormal{diam}(\mathcal{X}\times \Theta)\sqrt{\pi(d_x+d_{\theta})},\\
        &\delta_{\min} = \min_{t \in \mathcal{T}}\Bigg\{\min_{i,j \in \{1,\dots,m\}: i \neq j}\bigg\{\frac{1}{T_{i,t-1}} D_{i,\pi_1^t} (\theta_i,x_{i,0} || \theta_j,x_{j,0}) : |g_{i,t} - g_{j,t}| \geq \frac{\Delta_{\min}}{2m} \bigg\}\Bigg\},
    \end{align}
    $\mathcal{S}$ is the set of super-arms, $d_x$ is the dimensionality of $\mathcal{X}$, and $d_{\theta}$ is the dimensionality of $\Theta$.
    %chosen from the set of arms $\{1,\dots,m\}$ and $|\mathcal{S}|=2^m$.
\end{theorem}

We prove Theorem~{\ref{thm:klcucb-regret}} via two propositions. The first one bounds the regret increment per bad round. Assume the system is in state $\Bar{\bx}_t=(\Bar{x}_{1,t}, \Bar{x}_{2,t}, \dots, \Bar{x}_{m,t})$ in round $t$. If $t$ is a bad round, the super-arm selected by policy $\Tilde{\pi}$ will yield a lower expected reward than the optimal super-arm. The suboptimality gap is given by 
    \begin{align*}
        \Delta_t &= g\big(\btheta, f(\bar{\bx}_{t}, \pi^*_t)\big) - g\big(\btheta, f(\bar{\bx}_{t}, \Tilde{\pi}_t)\big)
    \end{align*}

\begin{proposition}[Regret increment per bad round] \label{prop:bad-round-increment}
    Given the assumption that set $\mathcal{X} \times \Theta$ is compact and that $g$ is locally $L_g$-Lipschitz continuous with respect to $x$, $\theta$ on $\mathcal{X} \times \Theta$, the regret incurred per bad round $t$, $\Delta_t$, is at most:
    \begin{align}
        \Delta_t \leq C L_g \textnormal{diam}(\mathcal{X} \times \Theta) \label{eq:regret-increment}
    \end{align}
\end{proposition}

The proof consists of using Assumptions~\ref{ass:log-lipschitz-cont}, \ref{ass:sub-gaussian}, and \ref{ass:dynamic-lipschitz} and extending known MAB results \citep{mintz2020nonstationary} to a combinatorial MAB setting by multiplying their upper bound for the regret increment per bad round by the number of arms since each super-arm can contain up to $C$ arms. Now that we have an upper bound on the regret increment per bad round, we proceed to bound the expected number of bad rounds for each super-arm $S \in \mathcal{S}$.

\begin{proposition}[Expected number of bad rounds] \label{prop-number-bad-rounds}
    Let $T^B(n)$ denote the number of bad rounds per super-arm for $n$ rounds of play using the COBRAH algorithm then:
    \begin{align}
        \E\big[T^B(n)\big] \leq \frac{4\big(B(\lceil \frac{m}{C}\rceil^{-4})\big)^2\log(n)}{\delta_{\min}^2} +  \frac{m^2\pi^2}{3} \label{eq:exp-num-bad-rounds}
    \end{align}
\end{proposition}

We provide a sketch of the proof. We bound the expected number of bad rounds separately based on whether the upper confidence bounds hold for all arms. For the case where all UCBs hold, we determine a threshold for the number of bad rounds that can occur which is a function of $n$ and the closest mean reward between any two arms. This means that a bad round will not occur if the UCBs hold and there has been sufficient exploration to detect a difference between the expected rewards for different arms. For the case where not all UCBs hold, we use the union of two events: the case where at least one UCB for an arm $i$ in a set of suboptimal super-arms is overestimated and where at least one UCB for an arm $j$ in a set of optimal super-arms is underestimated. Both of these events happen with a bounded probability leading to the desired result.

With these two intermediate results, we are ready to prove Theorem~\ref{thm:klcucb-regret} by multiplying the maximum regret increment per bad round, the number of super-arms, and the expected number of bad rounds per super-arm.

\subsection{Full Bandit Feedback}
\label{sec:policies-ff}

 We now consider the case where the provider rewards at each period $t$ are a nonlinear function of the expectation vector of all arms $\bg_t=(g_{1,t}, \dots, g_{m,t})$. In contrast with Section~\ref{sec:kl-cucb-sb-alg}, the rewards of all arms are observed at every period. Let $r_{\bg_t}(S)$ be the expected reward of playing super-arm $S$ at time $t$. We make two mild assumptions on the total expected reward $r_{\bg_t}(S)$:

\begin{assumption}[Monotonicity] \label{ass:monotonicity}
    Let $\bg_t$ be a vector representing the expected rewards of all arms at time $t$, i.e., $\bg_t=(g_{1,t}, \dots, g_{m,t})$. For each period $t \in \{1,\dots,n\}$, if $g_{i,t} \leq g'_{i,t}$ for all $i \in \{1,\dots,m\}$, then $r_{\bg_t}(S) \leq r_{\bg'_t}(S)$ for all $S \in \mathcal{S}_t$.
\end{assumption}
This assumption indicates that if a set of actions increases the rewards for all agents, then it must increase the global objective. Including this assumption introduces a partial order across all super arms allowing for proper dominance relationships and for a well defined optimal super arm set. Moreover, this assumption is met by many real world modeling scenarios. Consider behavioral and CHW interventions and where if an intervention improves the health of each individual in the cohort, the decision makers should benefit more.
\begin{assumption}[Bounded smoothness] \label{ass:bounded-smoothness}
    There exists a strictly increasing bounded smoothness function $f(\cdot)$ such that, for any two expectation vectors $\bg_t$ and $\bg'_t$ and period $t \in \{1,\dots,n\}$, $|r_{\bg_t}(S) - r_{\bg'_t}(S)| \leq f(\Lambda)$ if $\max_{i \in S}|g_{i,t} - g'_{i,t}| \leq \Lambda$.
\end{assumption}
This assumption can be seen as a generalization of the local Lipschitz assumptions placed on each arm. Likewise, it ensures that if the individual rewards obtained for each agent are close to each other, then the global objective rewards of the super-arms must also be close in value. 

\subsubsection{COBRAH-FF Algorithm}
\label{sec:kl-cucb-ff-alg}
The COBRAH Full Feedback (FF) algorithm pseudo-code is presented in Algorithm \ref{alg:kl-cucb-ff}. The steps themselves are similar to the COBRA-SB algorithm in terms of the initialization and UCB calculation steps. The key difference is in the super-arm selection. At each time point, the UCBs are used to solve a nonlinear optimization problem for maximizing $r_{\bg_t}$ over the set of all super-arms. Then, since observations are made for all agents (even those who were not selected on a given period), the parameters of all arms are updated.
\begin{algorithm2e}[ht]
    \linespread{1.5}\selectfont
    % \DontPrintSemicolon
    % \KwData{$(\by, \bz, \bar{\bb}, S, B, G, R)$}
    % \KwResult{$w^*$, $x^*$ = ($\hat{\bb}^*$, $\bs^*$, $\btheta^*$, $p^*$, $\mu^*$, $\alpha^*$, $\theta_0^*$, $\lambda^*$, $s_0^*$, $\beta^*$, $\gamma^*$, $\rho^*$)}
    \Begin{
        % Initialization rounds
        \For{$i \in \{1,\dots,m\}$}{
            $t\xleftarrow{}i$
            
            Play each $i$ once by choosing an arbitrary super-arm $S$ such that $i \in S$

            Update the number of times each arm $i$ has been played ($T_i$) and the vector of reward observations $\br_i = (r_{i,0}, \dots, r_{i,t})$ % z_{i,T_i}
        }
        \While{$t \leq n$}{
            $t \xleftarrow{} t+1$
            
            \For{$i \in \{1,\dots,m\}$}{
                Compute:
                
                $\hat{\theta}_i, \hat{x}_{i,0} = \argmin \Bigg\{- \sum\limits_{t \in \mathcal{T}_i} \log p(r_{i,t} | \theta_i, x_{i,t}) : x_{i,t+1} = f(\theta_i, x_{i,t}, y_{i,t}) \: \forall t \in \{0,\dots,n-1\} \Bigg\}$

                $g^{\textnormal{UCB}}_{i,t} = \max\limits_{\theta_i, x_{i,0} \in \Theta \times \mathcal{X}} \Big\{g\big(\theta_i, f_i^t(x_{i,0})\big) : \frac{1}{n(\mathcal{T}_i)} D_{i,\pi_1^t} \big(\theta_i, x_{i,0} || \hat{\theta}_i, \hat{x}_{i,0}\big) \leq B(t^{-4})\sqrt{\frac{4\log(t)}{T_{i,t-1}}} \Big\}$
            }
            Play $\Bar{S}_t \in \argmax\limits_{S \in \mathcal{S}} \big\{r_{\bg_t^{\textnormal{UCB}}}(S)\big\}$; i.e., set $y_{i,t}=1 \:\forall i \in \Bar{S}_t$ and $y_{i,t}=0 \:\forall i \notin \Bar{S}_t$

            Update $T_i \: \forall i \in \Bar{S}_t$ and $r_{i,t} \: \forall i \in \{1,\dots,m\}$
        }
    
    }
    \caption{COBRAH-FF algorithm. \label{alg:kl-cucb-ff}}
\end{algorithm2e}

\subsubsection{COBRAH-FF Regret}

Let $\Bar{\pi}=\pi_{1}^{n}$ be the complete policy given by the COBRAH-FF algorithm, $\pi^*$ denote the optimal action sequence obtained by a perfect oracle, $S_t^*$ be a super-arm in the set of optimal super-arms in period $t$, $\mathcal{S}_t^*$, and $\bg_t$ denote the true expected reward vector at time $t$. Because the rewards of all arms (including those that haven't been played) factor into the total provider reward at each period, the regret of the algorithm is given by:
\begin{align*}
    %R_{\Bar{\pi}}(n) &= \sum\limits_{t=1}^n \sum\limits_{i=1}^m g(\theta_i, f_{\pi^*}^t(x_{i,0})) -  \sum\limits_{t=1}^n \sum\limits_{i=1}^m g(\theta_i, f_{\Bar{\pi}}^t(x_{i,0}))\\
    R_{\Bar{\pi}}(n) &= \sum\limits_{t=1}^n \Big(r_{\bg_t}(S_t^*)-r_{\bg_t}(\Bar{S}_t)\Big),
\end{align*}
where $\Bar{S}_t$ is the super-arm selected by policy $\Bar{\pi}$ in period $t$.

\begin{corollary} \label{cor:regret-general}
    If the outcomes of the arms are independent, the regret of policy $\Bar{\pi}$ obtained using the COBRAH-FF algorithm is bounded by
    \begin{align}
        \E[R_{\Bar{\pi}}(n)] \leq C L_g \textnormal{diam}(\Theta \times \mathcal{X})|\mathcal{S}|\bigg(\frac{4\big(B(\lceil \frac{m}{C} \rceil)^{-4}\big)^2\log(n)}{\Bar{\delta}_{\min}^2} + \frac{m^2\pi^2}{3}\bigg),
    \end{align}
    where
    \begin{align}
        &\bar{\delta}_{\min} = \min_{t \in \mathcal{T}} \Bigg\{\min_{S, S' \in \mathcal{S}}\bigg\{\min_{i \in S, j \in S'}\Big\{D_{i,\pi_1^t} (\theta_i,x_{i,0} || \theta_j,x_{j,0}) : \big|\bg_{t}(S) - \bg_{t}(S')\big| \geq \tfrac{1}{2}f^{-1}(\Bar{\Delta}_{\min})\Big\} \bigg\}\Bigg\},\\
        &\Bar{\Delta}_{\min} = \min_{i \in 1,\dots,m}\bigg\{\min_{t \in \mathcal{T}} \Big\{r_{\bg_t}(S_t^*)-\max_{S_t \in \mathcal{S}_t^B: i\in S_t} r_{\bg_t}(S_t)\Big\}\bigg\}\\
    \end{align}
\end{corollary}

With the increment per bad round being the same as in the semi-bandit case, the proof of the expected number of bad rounds follows similar steps to the proof of Proposition~\ref{prop-number-bad-rounds}. Due to the possibility of nonlinear rewards functions, Assumptions~\ref{ass:monotonicity} and \ref{ass:bounded-smoothness} are used to relate the expected rewards from choosing a suboptimal super-arm to an optimal super-arm.

%%%%%% NUMERICAL EXPERIMENTS %%%%%%
\section{Numerical Experiments}
\label{sec:experiments}

In this section, we describe our numerical experiments that compare one of our policies to existing approaches in the literature. We focus on a policy computed with the COBRAH-SB algorithm, where rewards are observed only for the chosen arms. In Section~\ref{sec:exp-tuned-klcucb}, we present a tuned version of our COBRAH algorithm under semi-bandit feedback that leads to faster convergence in practice. In Section~\ref{sec:exp-compar-algs}, we describe the comparison algorithms used in our experiments. In Section~\ref{sec:exp-sim-setup}, we discuss details of our simulation setup. Finally, in Section~\ref{sec:exp-results}, we present our experimental results. 

\subsection{Tuned COBRAH Algorithms}
\label{sec:exp-tuned-klcucb}

It is well established that the high probability bounds derived in UCB algorithms are often too conservative \citep{auer2002finite, garivier2008upper, bouneffouf2016multi}. To focus on the empirical analysis of the COBRAH-SB and COBRAH-FF algorithms, we developed tuned versions which are shown in Algorithms~\ref{alg:tuned-kl-cucb-sb} and \ref{alg:tuned-kl-cucb-ff}. 

% Tuned KL-CUCB - Semi-Bandit
\begin{algorithm2e}[ht]
    \linespread{1.5}\selectfont
    % \DontPrintSemicolon
    % \KwData{$(\by, \bz, \bar{\bb}, S, B, G, R)$}
    % \KwResult{$w^*$, $x^*$ = ($\hat{\bb}^*$, $\bs^*$, $\btheta^*$, $p^*$, $\mu^*$, $\alpha^*$, $\theta_0^*$, $\lambda^*$, $s_0^*$, $\beta^*$, $\gamma^*$, $\rho^*$)}
    \Begin{
        % Initialization rounds
        \For{$i \in \{1,\dots,m\}$}{
            $t\xleftarrow{}i$
            
            Play each $i$ once by choosing an arbitrary super-arm $S$ such that $i \in S$

            Update the number of times each arm $i$ has been played ($T_i$) and the vector of reward observations $\br_i = (r_{i,0}, \dots, r_{i,t})$ % z_{i,T_i}
        }
        \While{$t \leq n$}{
            $t \xleftarrow{} t+1$
            
            \For{$i \in \{1,\dots,m\}$}{
                Compute:
                
                $\hat{\theta}_i, \hat{x}_{i,0} = \argmin \Bigg\{- \sum\limits_{t \in \mathcal{T}_i} \log p(r_{i,t} | \theta_i, x_{i,t}) : x_{i,t+1} = f(\theta_i, x_{i,t}, y_{i,t}) \: \forall t \in \{0,\dots,n-1\} \Bigg\}$

                $g^{\textnormal{UCB}}_{i,t} = \max\limits_{\theta_i, x_{i,0} \in \Theta \times \mathcal{X}} \bigg\{g(\theta_i, f_i^t(x_{i,0})) : \frac{1}{n(\mathcal{T}_i)} D_{i,\pi_1^t} \big(\theta_i, x_{i,0} || \hat{\theta}_i, \hat{x}_{i,0}\big) \leq \sqrt{\min\big\{\frac{\eta}{4}, \mathcal{V}_{i,\pi_1^T}(\theta_i, x_{i,0}||\hat{\theta}_i, \hat{x}_{i,0})\big\}\frac{\log(t)}{T_{i,t-1}}} \bigg\}$
            }
            Play $S_t = \argmax\limits_{S \in \mathcal{S}} \sum_{i \in S} g_{i,t}^{\textnormal{UCB}}$; i.e., set $y_{i,t}=1 \:\forall i \in S_t$ and $y_{i,t}=0 \:\forall i \notin S_t$

            Update $T_i \: \forall i \in S_t$ and $r_{i,t} \: \forall i \in \{1,\dots,m\}$
        }
    
    }
    \caption{Tuned COBRAH-SB algorithm. \label{alg:tuned-kl-cucb-sb}}
\end{algorithm2e}

% Comment that UCB bounds derived theoretically are often too conservative

% If MLE are consistent and in the interior of the feasible region, they are asymptotically normally distributed with variance equal to their Fisher information

Similarly to \cite{auer2002finite}, we used asymptotic confidence intervals in the tuned variants of our COBRAH-SB and COBRAH-FF algorithms. To obtain asymptotic bounds that offer faster convergence, we note that if MLE estimates are consistent and in the interior of the feasible region, they are asymptotically normally distributed with variance equal to their Fisher information \citep{shapiro1993asymptotic}. We then use the the delta method \citep{keener2010theoretical} to derive the asymptotic variance of the trajectory KL divergence, which we denote by $\mathcal{V}_{i,\pi_1^T}(\theta_i, x_{i,0}||\hat{\theta}_i, \hat{x}_{i,0})$.

% \subsection{Simulation Setup}
% \label{sec:exp-sim-setup}

% \ka{UPDATE} Experiments were run using a planning horizon of $N=520$ periods (10 years of weekly planning periods), 2000 patients, and a capacity of 100 visits/period. Simulations were run using HTCondor the Center for High Throughput Computing at the University of Wisconsin-Madison. We used Python 3.12 to run the experiments. We solve 10 replications of each instance using different random seeds. 

% \subsection{Measurement Noise}
% \label{sec:exp-noise}

% We assume FBG measurement noise ($\xi_t$) follows a Laplace distribution with mean zero. The variance of $\xi_t$ was estimated using data obtained from our partner company, NanoHealth. Regarding the benefit function, we assume that it is Bernoulli with probability of success $\frac{z_{t-1}+y_t-z_{t-1}y_t}{1+\exp(-B_t)}$.

\subsection{Comparison Algorithms}
\label{sec:exp-compar-algs}

We assess the performance of our COBRAH-SB and COBRAH-FF algorithms against three baseline algorithms, with the results shown in Sections~\ref{sec:exp-results} and \ref{sec:exp-results-case-study}, respectively. The first baseline is based on a random allocation policy, while the other two are approaches from the literature.

% \subsubsection{Random Policy}

% As a baseline, we utilize a policy that randomly chooses which participants to allocate appointments to at each period.

% \subsubsection{Enrollment Algorithm}

% \ka{Will this be included?} The Enrollment Algorithm (EA) is an approximate dynamic programming approach that was developed for planning CHW interventions in LMICs under full information \citep{adams2023planning}. To compare our approach to theirs, we adapted it their algorithm to the partial information setting.

\begin{enumerate}
    \item \textbf{Random Policy:} As a baseline, we utilize a policy that chooses which agents to allocate resources to uniformly at random at each period.
    \item  \textbf{Combinatorial UCB: } Next, we consider the Combinatorial UCB (CUCB) algorithm developed by \cite{chen2013combinatorial}. Similarly to our problem, the authors focus on a class of combinatorial MAB problems where simple arms with unknown reward distributions compose super arms. We use their original semi-bandit approach for comparison against COBRAH-SB and adapt their approach to the full bandit case for comparison with COBRAH-FF.
    \item \textbf{Combinatorial UCB with Sliding Window: }Finally, we consider the Combinatorial UCB algorithm with sliding window (SW-UCB) developed by \cite{chen2021combinatorial}. This approach is similar to the previous one \citep{chen2013combinatorial} except for the addition of a sliding window to the UCB computations. This is meant to capture problems with nonstationary reward distributions. We choose the window size $w$ for the distribution-independent case, calculated with the following formula recommended by the authors, $w = \min\big\{m^{1/3} T^{2/3} K^{-1/3} V^{-2/3}, T\big\}$, where $m$ represents the number of arms, $T$ represents the number of planning periods, $K$ is the maximum number of arms that can be triggered by an action in any round, and $V$ is the variation given by the formula $V=\sum_{t=2}^T \|\bmu_t - \bmu_{t-1}\|_{\infty}$. For our assumptions, $V$ is proportional to $T$, leading to a window size of $w = \min\bigg\{\sqrt[3]{\frac{m}{C}}, T\bigg\} $.
\end{enumerate}

\subsection{Simulation Setup}
\label{sec:exp-sim-setup}

In this section, we provide details on the COBRAH instantiation and on the setup for the experiments with synthetic data. Experiments were run using an Apple M2 chip and 24 GB of RAM.

\subsubsection{Combinatorial ROGUE Bandit Instantiation}

Using the setting of Section \ref{sec:model-framework} we consider a generalized linear model (GLM) setting with Bernoulli rewards. For each arm we set $x_i \in [0,1]^2$ and $\theta_i \in [0,1]$. The distribution fo the reward of each agent is modeled with the following logistic link function:
\begin{align}
    g(x_{i,t},\theta_i) &= \frac{1}{1 + \exp\big(-(\nu_i^\top\theta_i + \omega_i^\top x_{i,t})\big)}
\end{align}
Where $\nu = [1], \omega = [1, -1]^\top$. The system dynamics for each arm are assumed to be piecewise linear with the form $\bx_{i,t+1} =  \text{proj}_{[0,1]} \{D \bx_{i,t} + Q y_{i,t} + K\}$, where $\text{proj}_{[0,1]}$ is the operator that projects the state back onto the closed interval $[0,1]$, and
\begin{align}
    D =
    \begin{bmatrix}
        d_1 & 0\\
        0 & d_2
    \end{bmatrix},\:
    Q = 
    \begin{bmatrix}
        q_1 \\
        q_2
    \end{bmatrix},\:
    K =
    \begin{bmatrix}
        k_1 \\
        k_2
    \end{bmatrix}.
\end{align}
%In the full feedback case, we also restrict enrollment to patients who were previously enrolled or screened by multiplying the reward generated with a Binomial distribution by $(z_{i,t-1} + y_{i,t} - z_{i,t-1} y_{i,t})$.

The state dynamics are set so that the different components of $x_i$ evolve independently.

% \ka{Discuss disturbance?}

\subsubsection{Numerical Parameters}

For the semi-bandit experiments with synthetic data, we chose a planning horizon of 4000 periods, 100 arms, a budget of 20  per period, and 20 replications of each instance. The parameters for the matrices associated with the dynamics, the patient parameters $\theta_i$, and initial states $x_{i,0}$ were generated by sampling from uniform distributions that are defined over intervals with the lower and upper values shown in Table~\ref{tab:params-glm}.
\begin{table}[ht]
\centering
\begin{tabular}{c|cc}
    Parameter & Lower value & Upper value \\\hline
    $d_1$, $d_2$ & 0.5 & 1.0\\
    $q'_1$, $q'_2$ & 0.1 & 2.0\\
    $k_1$, $k_2$ & 0.1 & 2.0\\
    $b$, $a$ & 0.0 & 1.0\\
    $\theta$ & 0.0 & 1.0
\end{tabular}
\caption{Lower and upper values of intervals used to sample parameters using uniform distributions} \label{tab:params-glm}
\end{table}
Note that $q_1=-q'_1 - k_1$, $q_2 = q'_2 - k_2$, and that states are limited to take on values between 0 and 1, therefore any values lower than 0 are modified to 0 and any values greater than 1 are modified to 1.

%The purpose of these experiments is to verify the performance and scalability of an approach that may be used for urban settings 

\subsection{Experimental Results}
\label{sec:exp-results}

\begin{figure}[ht]
\centering
\subfloat[Cumulative regret\label{fig:cum-regret-sb}]{\includegraphics[width=0.51\textwidth, trim={0cm 0 0cm 0cm},clip]{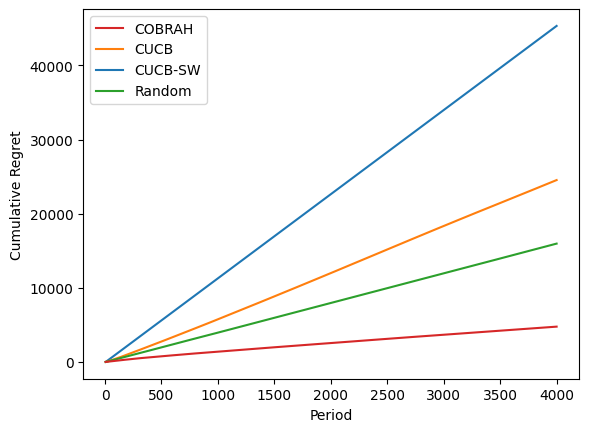}}
\subfloat[Longrun average reward and its rolling average with 200-period window \label{fig:avg-reward-sb}]{\includegraphics[width=0.49\textwidth, trim={0cm 0 0cm 0cm},clip]{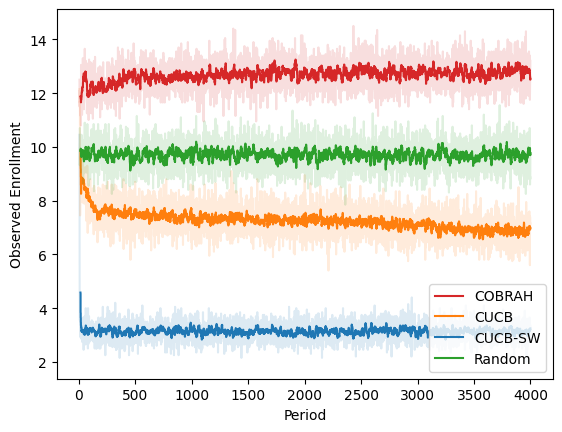}}
\vspace{5pt}
\caption{Results for semi-bandit version of algorithms after 4000 rounds of play with 100 arms and 20\% capacity.}
\end{figure}

 Figure~\ref{fig:cum-regret-sb} shows the expected cumulative regret comparison for COBRAH-SB, CUCB, and SW-UCB (with the latter two defined in Section~\ref{sec:exp-compar-algs}). The results show that the cumulative regret increases at a slower rate for the COBRAH-SB algorithm than for the comparison algorithms, indicating that it learns how to maximize rewards faster, leading to lower cumulative regret. Furthermore, CUCB outperforms SW-UCB, suggesting that using only the latest portion of the state and action history is detrimental to learning in the semi-bandit case -- even though this is a nonstationary setting.

Figure~\ref{fig:avg-reward-sb} compares the long-run  reward of the algorithms.
%which is calculated by subtracting the enrollment probability with and without an appointment allocated to a participant. Then, the net probabilities are averaged over all periods up to each period. Note that the net reward may take on negative values if the enrollment probability is greater without an appointment than with it.
The plot shows that the COBRAH-SB algorithm maintains high long-run  average rewards (above 12) throughout the entire horizon, while the comparison algorithms quickly drop to around 1 for SW-UCB and 7 for CUCB. These results suggest that resources are allocated more efficiently by our algorithm. If we think of this results as replicating a behavioral intervention, this would translate to higher enrollment in the long run, benefiting the central decision maker. We provide additional discussions on the managerial implications of our algorithms in Section~\ref{sec:exp-results-managerial-imp}.

%%%%%% CASE STUDY %%%%%%
\section{Case Study}
\label{sec:case-study}

In this section, we go over details of our case study using observational data. We collaborated with and obtained data from NanoHealth, a social enterprise based in Hyderabad (State of Telangana), India \citep{NanoHealth2020}. NanoHealth employs a team of CHWs to operate a diabetes treatment program in the fourth largest city in India, with a projected population of over 10 million \citep{markandey2023changing, census:2011}. Each CHW is equipped with a ``Doc-in-the-Bag” kit and tablet that allows them to obtain and record certain anthropomorphic measurements and vitals \citep{deo2021community}. 
%Screening visits are provided to identify patients with diabetes who were undiagnosed, while management visits are provided to patients who are screened, identified as high risk based on the anthropometric measurements collected, and decide to enroll in NanoHealth's treatment program \citep{boutilier2021risk}. 
The dataset used for our case study experiments was collected between 2015 and 2018, with each CHW conducting an average of 144 visits per month ($\approx$5 per day). More details on this dataset are provided in Appendix~\ref{app:nanohealth-data}. The remainder of the section is organized as follows: Section~\ref{sec:case-study-context} describes the study context, Section~\ref{sec:case-study-exp} details the experimental setup, and Section~\ref{sec:exp-results-case-study} presents the case study results.

\subsection{Case Study Context}
\label{sec:case-study-context}

We provide some context that is relevant to understand the modeling approach used for planning a targeted CHW intervention for diabetes care in low- and middle-income countries (LMICs).

\subsubsection{Diabetes Overview}

We focus on type 2 diabetes, which comprises over 90\% of the overall diabetic population. Diabetes is a noncommunicable and chronic disease that requires an early diagnosis and continued care to reduce the likelihood that it will lead to morbidity and mortality. Early symptoms of diabetes may include frequent urination and excessive fatigue, thirst, and hunger. If it goes undetected and unmanaged, long-term high blood glucose levels may cause damage to the blood vessels, leading to a vast array of complications, such as myocardial infarctions, strokes, kidney failure, limb amputations, retinopathy, and neuropathy \citep{idf2021atlas}. Diabetes can be diagnosed using several tests, including random blood glucose, post-prandial blood glucose, fasting blood glucose (FBG), oral glucose tolerance test, and hemoglobin A1c (HbA1c). %Considering the tradeoff of practical feasibility and diagnostic accuracy \citep{hoyer2018utility}, we utilize FBG measurements as side information when estimating program enrollment benefit. \ka{(UPDATE?)}

To underscore the importance of screening to tackle the global diabetes epidemic, we provide some background on the pathophysiology of diabetes. High blood glucose levels are caused by two primary mechanisms. First, insulin resistance, that is, the ineffective use of insulin (the ``key" that allows glucose to enter cells), leading to glucose accumulation in the bloodstream. Second, a decline in the function of beta cells (i.e., insulin-producing cells in the pancreas), which occurs due to excessive insulin secretion by the body for an extended time to maintain normal blood glucose in the presence of insulin resistance. It is estimated that by the time diabetes is diagnosed, 40-50\% of beta-cell function has already been lost \citep{wysham2020beta}. Inconspicuous early symptoms and advanced progression by the time of diagnosis illustrate the need for screening to allow for early detection of diabetes, a measure that would likely offset the costs to manage it.

\subsubsection{Economic Impact of Diabetes in LMICs}

High-income countries contributed the most to the global diabetes burden at US\$804.36 billion, representing 1.2\% of their average gross domestic product (GDP). Meanwhile, low-income countries had a relative economic burden of 0.7\% of their average GDP and, middle-income countries, 1.8\%, representing the highest relative cost among all country income groups \citep{bommer2017global}. Nearly 60\% of direct global diabetes costs were borne by LMICs, where treatment costs are often paid out-of-pocket \citep{seuring2015economic}, leading families to face catastrophic health expenditures \citep{sathyanath2022economic}. 

An estimated 34.7\% of the global burden was attributed to indirect costs, defined as costs due to premature mortality and morbidity \citep{bommer2017global, idf2015atlas}. Indirect costs due to morbidity occur in the form of lost workdays due to sickness (absenteeism), reduced productivity while working (presenteeism), and labor market dropout due to disability \citep{pradeepa2017prevalence, liu2023projected}. When examining the composition of indirect costs of diabetes across economic regions, \cite{bommer2017global} note that morbidity-related factors dominate in high-income countries (64.5\%), while premature mortality is the greatest cost component in middle-income (63.6\%) and low-income (90.6\%) countries. Because the labor market effects of undiagnosed diabetes are unclear \citep{minor2016comparison}, existing estimates are based on the conservative assumption that undiagnosed diabetes incurred no labor costs, suggesting that these costs may be much higher in countries with a high prevalence of undiagnosed diabetes.

\subsubsection{Strategies for Cost-Effective Diabetes Care in LMICs}

Diabetes places a heavy burden on health systems, a challenging issue for LMICs due to health system limitations such as workforce shortages and high economic costs of hospitalization, laboratory tests, and pharmaceutical interventions. 
%Additionally, diabetes has a significant economic burden when considering both direct (e.g., ) and indirect costs (e.g., loss productivity, absenteeism, job loss, disability assistance) \citep{pradeepa2017prevalence, liu2023projected}. 
%Among strategies to increase access to care in this context, employing CHWs and relying on public-private partnerships (PPPs) present viable pathways. 
CHW programs provide an affordable and effective solution to screen and treat patients with diabetes \citep{alaofe2017community, jeet2017community, gyawali2021effectiveness}. Rather than relying on existing health systems, CHW programs employ non-medical personnel and expand health system capacity through task shifting by training members of local communities to perform screening tests and provide culturally tailored education on diabetes management. 

%Diagnosing patients through a blood test to obtain a HbA1c measurement is the standard method in high-income countries, but it is not always feasible in LMICs due to its high cost. Measuring the FBG is an alternative that is cheaper and can be done by the patients themselves or by a trained non-medical professional. Upon diagnosis, patients with T2D can receive pharmacological care (i.e., being prescribed medication) and diabetes education to understand how to manage their condition, including learning healthier lifestyle choices. While performing blood tests and prescribing medication requires medical professionals, lab technicians, and appropriate infrastructure, other tasks could be assigned to non-medical personnel (e.g., CHWs) with appropriate training.

Furthermore, Public-Private Partnerships (PPPs) provide a viable option for scaling and implementing CHW interventions. These partnerships between governments and businesses are a viable path to address the diabetes epidemic in LMICs for multiple reasons. First, businesses that are part of PPPs can provide capital to finance these programs. For businesses, working with CHW organizations can provide a cost-effective approach for early detection and management of diabetes complications, driving a healthier and more productive workforce \citep{wnuk2023workplace, penalvo2021effectiveness, groeneveld2010lifestyle}. Second, PPPs allow for targeting populations that are at a high risk of diabetes. Researchers have identified unhealthy diets and sedentary lifestyles as key drivers of the global diabetes epidemic \citep{zheng2018global}, with office workers being particularly susceptible and affected by type 2 diabetes \citep{parry2013contribution, holla2022risk, ryu2016office}. Yet, being an office worker or having a diabetes diagnosis do not necessarily lead to greater knowledge of diabetes risk factors \citep{dika2023practice, saeedi2020cardiovascular}, showing the need to educate this population on diabetes prevention and management. Third, businesses can provide the space required for these interventions to take place, as LMICs currently have insufficient health infrastructure to tend to the needs of their populations \citep{phelan2022challenges, bollyky2017lower}. While previous studies have shown low cost-effectiveness for population-level diabetes screening in LMICs for all screening frequencies, these  programs became more cost-effective if combined with treatment provision or if they targeted at-risk individuals, underlining the need for targeted interventions and care provision models that can accommodate not only sporadic screening, but also continuous care \citep{kaur2022cost, toscano2015cost}. Finally, we note that our appointment allocation framework addresses a key challenge of successful PPP and CHW intervention implementation, which lies in programming, accountability, and performance assessment \citep{fanelli2020insights, kok2021community}. 

\subsection{Case Study Experiments}
\label{sec:case-study-exp}

In our case study, a provider seeks to build worker wellness plans guiding CHW appointment allocation decisions to prevent, diagnose, and treat diabetes among their workforce. 
%Screening allows workers who have diabetes and are unaware to get diagnosed. Those who are screened and found to have a high FBG (hyperglycemia) are recommended to enroll in treatment to receive follow-up appointments and improve glycemic control (i.e., lower FBG below a given threshold). 
A limited number of appointments is available per planning period depending on the number of CHWs employed, therefore the provider must carefully choose which workers to see at each period to screen and enroll new workers in the wellness program, provide education to enrolled participants, and collect data to improve decision-making. Because the motivational state of each participant depends on their diabetes status and stage, current knowledge on diabetes prevention and management, perception of self-efficacy to prevent or manage it, among other factors, allocation decisions must be able to capture trade-offs that affect enrollment in the wellness program at the individual level.
%We assume that the provider only observes the blood glucose for those seen in appointments and that the planning problem takes place prior to any enrollment.

Due to the possibility of allocating appointments for multiple workers per period and to the rewards being nonstationary, the provider problem falls within combinatorial and nonstationary bandit frameworks and can be modeled using the combinatorial nonstationary bandit framework described in Section~\ref{sec:comb-ns-prob}. In particular, the set of workers being targeted corresponds to the set of arms, the motivational level to enroll/stay enrolled in the program corresponds to the state of each arm, and the payoffs (rewards) correspond to the number of participants enrolled in the program at each time period. Note that states evolve whether a worker has an appointment allocation or not, although these decisions may affect their state transitions. Rewards are nonstationary due to the dependence on the workers' underlying states, which are nonstationary. Our goal is to choose those to allocate appointments for at each time period (arm selections) such that program enrollment (total provider reward) is maximized. We assume that the rewards are stochastic and consider
%both the semi-bandit case (rewards are only observed for chosen arms) and 
the full feedback case (rewards are observed for all arms). 

\subsubsection{Case Study Framework}
\label{sec:case-study-framework}

%\ka{(EDIT HERE)} We assume that $x_{i,t}$ is composed of the log-FBG level ($b_{i,t}$), the adverse factors of enrollment ($s_{i,t}$), and the cognitive weight of adverse factors of enrollment ($c_{i,t}$). The product $s_{i,t} c_{i,t}$ reflects the resulting perceived adverse factors, and modeling them separately allows for $c_{i,t}$ to reduce over time as participants become more educated on diabetes management, even if $s_{i,t}$ increases with each management appointment. The provider accrues rewards for each participant-period of enrollment, which we denote by the binary quantity $z_{i,t}$. We assume rewards are bounded and governed by nonlinear dynamics (e.g., composition of nonlinear value function with a linear threshold). Specifically, a patient may enroll in treatment if the benefit of enrollment, $B_{i,t}$, is nonnegative and will drop out (or not enroll) in treatment otherwise. We incorporate random variables $\xi_{t}$ to represent the measurement noise in $b_{i,t}$, with $\xi_{t}$ assumed to have mean zero. 

We assume that $x_{i,t}$ is composed of beneficial enrollment factors ($b_{i,t}$) and adverse enrollment factors ($a_{i,t}$). This modeling choice seeks to incorporate factors related to a worker's utility, and, consequently, their enrollment choices. The provider accrues rewards for each patient-period of enrollment, which we denote by the binary quantity $z_{i,t}$. We assume rewards are bounded and governed by nonlinear dynamics (e.g., composition of nonlinear function with a linear threshold). Specifically, a patient may enroll in the program if its net advantage outweighs the disadvantage, that is, if $b_{i,t} \geq a_{i,t}$. Otherwise, we assume participants will drop out (or not enroll) in the wellness program. Our parameter set for each arm is represented by $\theta_i$. We model this problem as a dynamic logistic model \citep{filippi2010parametric} with rewards $z_{i,t}|\{\theta_i, x_{i,t}\}$ that follow a Bernoulli distribution with mean $g(\cdot)$, which can be interpreted as a link function of a generalized linear model (GLM), allowing for its use in a wide class of problems. For our application, we consider the specific case where it is a logit function, which satisfies Assumptions~\ref{ass:cond-independence} and \ref{ass:log-concave} since rewards are bounded and restricted to the interval $[0,1]$. States $x_{i,t}$ are assumed to evolve according to known dynamics of the form $x_{i,t+1} = f(\theta_i,x_{i,t}, y_{i,t})$.
We again consider piecewise linear dynamics similar
%to Appendix~\ref{app:param-est} for details on the parameter estimation procedure for this case study and 
to Section~\ref{sec:exp-sim-setup}. However, for this case study we impose further structure. Namely, we ensure that if a participant is visited this will increase their adverse factors while decreasing their beneficial factors, while if they are not visited the opposite will occur. This models the effects of habituation, cognitive burden, and stigma that affect participation in CHW interventions with too frequent a visit schedule.

\subsubsection{Case Study Algorithm}

Similarly to Algorithm~\ref{alg:tuned-kl-cucb-sb} shown in the numerical experiments with synthetic data, we developed a tuned version of our full feedback algorithm for faster convergence in practice (Algorithm~\ref{alg:tuned-kl-cucb-ff}).

% Tuned KL-CUCB - Full information
\begin{algorithm2e}[ht]
    \linespread{1.5}\selectfont
    % \DontPrintSemicolon
    % \KwData{$(\by, \bz, \bar{\bb}, S, B, G, R)$}
    % \KwResult{$w^*$, $x^*$ = ($\hat{\bb}^*$, $\bs^*$, $\btheta^*$, $p^*$, $\mu^*$, $\alpha^*$, $\theta_0^*$, $\lambda^*$, $s_0^*$, $\beta^*$, $\gamma^*$, $\rho^*$)}
    \Begin{
        % Initialization rounds
        \For{$i \in \{1,\dots,m\}$}{
            $t\xleftarrow{}i$
            
            Play each $i$ once by choosing an arbitrary super-arm $S$ such that $i \in S$

            Update the number of times each arm $i$ has been played ($T_i$) and the vector of reward observations $\br_i = (r_{i,0}, \dots, r_{i,t})$ % z_{i,T_i}
        }
        \While{$t \leq n$}{
            $t \xleftarrow{} t+1$
            
            \For{$i \in \{1,\dots,m\}$}{
                Compute:
                
                $\hat{\theta}_i, \hat{x}_{i,0} = \argmin \Bigg\{- \sum\limits_{t \in \mathcal{T}_i} \log p(r_{i,t} | \theta_i, x_{i,t}) : x_{i,t+1} = f(\theta_i, x_{i,t}, y_{i,t}) \: \forall t \in \{0,\dots,n-1\} \Bigg\}$
    
                $g^{\textnormal{UCB}}_{i,t} = \max\limits_{\theta_i, x_{i,0} \in \Theta \times \mathcal{X}} \bigg\{g(\theta_i, f_i^t(x_{i,0})) : \frac{1}{n(\mathcal{T}_i)} D_{i,\pi_1^t} \big(\theta_i, x_{i,0} || \hat{\theta}_i, \hat{x}_{i,0}\big) \leq$ \\
                $\hspace{7.25cm}\sqrt{\min\big\{\frac{\eta}{4}, \mathcal{V}_{i,\pi_1^T}(\theta_i, x_{i,0}||\hat{\theta}_i, \hat{x}_{i,0})\big\}\frac{\log(t)}{T_{i,t-1}}} \bigg\}$
            }
            Play $S_t = \argmax\limits_{S \in \mathcal{S}} \Big\{r_{\bg_t^{\textnormal{UCB}}}(S) \Big\}$; i.e., set $y_{i,t}=1 \:\forall i \in S_t$ and $y_{i,t}=0 \:\forall i \notin S_t$

            Update $T_i \: \forall i \in S_t$ and $r_{i,t} \: \forall i \in \{1,\dots,m\}$
        }
    
    }
    \caption{Tuned COBRAH-FF algorithm. \label{alg:tuned-kl-cucb-ff}}
\end{algorithm2e}

\subsubsection{Case Study Instantiation}
\label{sec:case-study-exp-setup}

Under the assumption that businesses can easily obtain information on the enrollment status of their workers participating in a CHW intervention such as the one described in Section~\ref{sec:case-study}, we utilize the full feedback algorithm presented in Section~\ref{sec:policies-ff} for our case study experiments. Prior to running our simulation, we used a grid search method to fit patient parameters to their historical data. We assume that each patient's initial beneficial and adverse effect states are all 0 ($x_{i,0} = [0 \ 0]^\top$) and $d_1, d_2, q_1, q_2, k_1, k_2, \theta$ are searched on $[0,1]$. We use the parameter combination that maximizes the likelihood of observing the historical enrollment decisions of each patient as their ground-truth parameters.

We performed several simulation studies to see how our COBRAH methods perform against existing approaches across various intervention budgets. For each simulation we used a planning horizon of 312 periods (corresponding to 6 years of weekly periods), 378 participants, budgets of 5\%, 10\%, 20\%, 40\%, and 80\% appointments per period (relative to the number of program participants), and 10 replications of each instance.

\subsection{Case Study Results}
\label{sec:exp-results-case-study}
\begin{figure}[ht]
\centering
\subfloat[Regret with 5\% capacity \label{fig:case-study-regret-5}]{\includegraphics[width=0.32\textwidth, trim={0 0cm 0.4cm 1.15cm},clip]{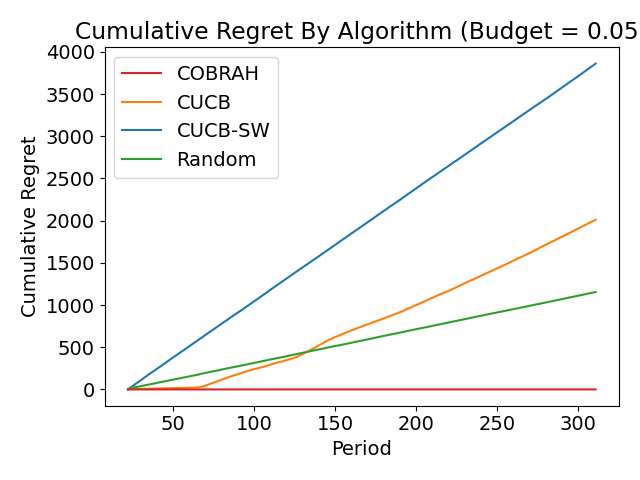}}
% \subfloat[Reward with 5\% capacity \label{fig:case-study-reward-5}]{\includegraphics[width=0.32\textwidth, trim={0 0.3cm 1.2cm 1.35cm},clip]{figures/reward_ROGUE_0.05.png}}
% \subfloat[Enrollment with 5\% capacity \label{fig:case-study-enrollment-5}]{\includegraphics[width=0.48\textwidth, trim={0 0.3cm 1.2cm 1.35cm},clip]{figures/enrollment_COBRAH_0.05.png}}\\[-1ex]
% \centering
\subfloat[Regret with 10\% capacity \label{fig:case-study-regret-10}]{\includegraphics[width=0.32\textwidth, trim={0 0cm 0.4cm 1.15cm},clip]{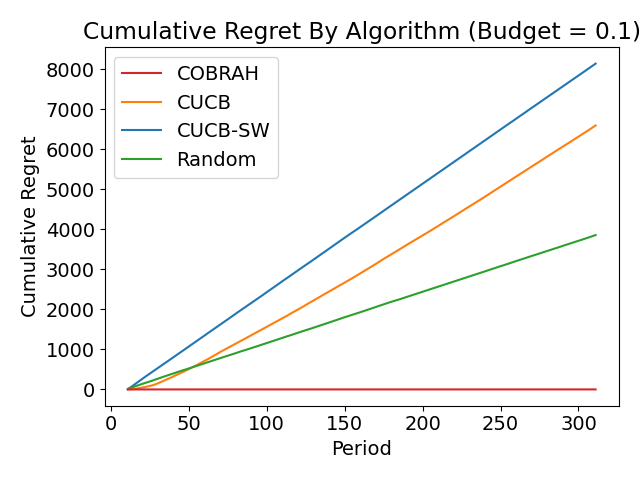}}
% \subfloat[Reward with 10\% capacity \label{fig:case-study-reward-10}]{\includegraphics[width=0.32\textwidth, trim={0 0.3cm 1.2cm 1.35cm},clip]{figures/reward_ROGUE_0.1.png}}
% \subfloat[Enrollment with 10\% capacity \label{fig:case-study-enrollment-10}]{\includegraphics[width=0.48\textwidth, trim={0 0.3cm 1.2cm 1.35cm},clip]{figures/enrollment_COBRAH_0.1.png}}\\[-1ex]
% \centering
\subfloat[Regret with 20\% capacity \label{fig:case-study-regret-20}]{\includegraphics[width=0.32\textwidth, trim={0 0cm 0.4cm 1.15cm},clip]{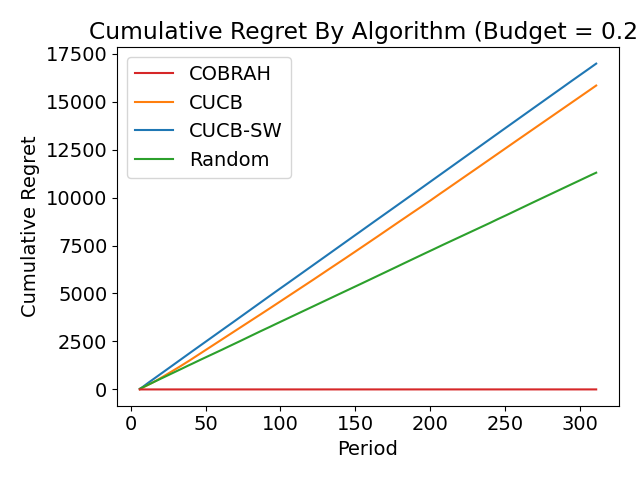}}\\[-1ex]
% \subfloat[Reward with 20\% capacity \label{fig:case-study-reward-20}]{\includegraphics[width=0.32\textwidth, trim={0 0.3cm 1.2cm 1.35cm},clip]{figures/reward_ROGUE_0.2.png}}
% \subfloat[Enrollment with 20\% capacity \label{fig:case-study-enrollment-20}]{\includegraphics[width=0.48\textwidth, trim={0 0.3cm 1.2cm 1.35cm},clip]{figures/enrollment_COBRAH_0.2.png}}\\[-1ex]
% \centering
\subfloat[Regret with 40\% capacity \label{fig:case-study-regret-40}]{\includegraphics[width=0.32\textwidth, trim={0 0cm 0.4cm 1.15cm},clip]{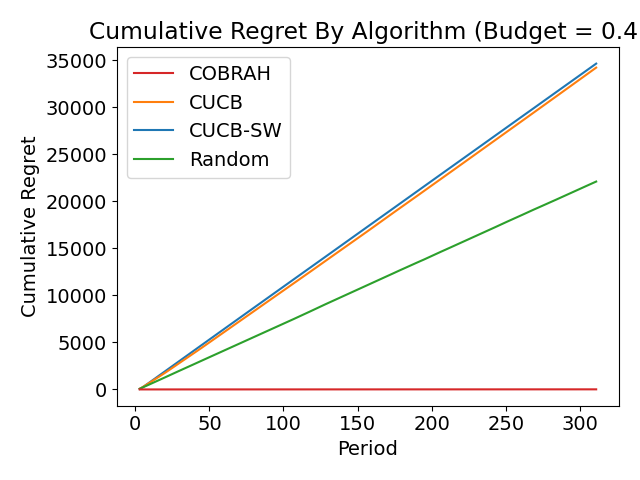}}
% \subfloat[Reward with 40\% capacity \label{fig:case-study-reward-40}]{\includegraphics[width=0.32\textwidth, trim={0 0.3cm 1.2cm 1.35cm},clip]{figures/reward_ROGUE_0.4.png}}
% \subfloat[Enrollment with 40\% capacity \label{fig:case-study-enrollment-40}]{\includegraphics[width=0.48\textwidth, trim={0 0.3cm 1.2cm 1.35cm},clip]{figures/enrollment_COBRAH_0.4.png}}\\[-1ex]
% \centering
\subfloat[Regret with 80\% capacity \label{fig:case-study-regret-80}]{\includegraphics[width=0.32\textwidth, trim={0 0cm 0.4cm 1.15cm},clip]{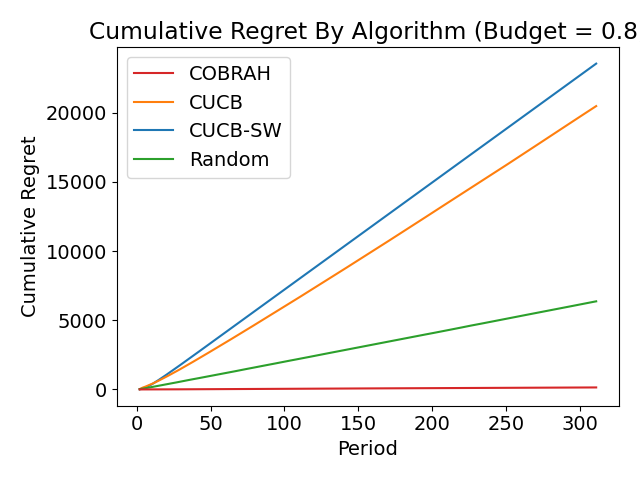}}\\[0.2cm]
% \subfloat[Reward with 60\% capacity \label{fig:case-study-reward-60}]{\includegraphics[width=0.32\textwidth, trim={0 0.3cm 1.2cm 1.35cm},clip]{figures/reward_ROGUE_0.6.png}}
% \subfloat[Enrollment with 80\% capacity \label{fig:case-study-enrollment-80}]{\includegraphics[width=0.48\textwidth, trim={0 0.3cm 1.2cm 1.35cm},clip]{figures/enrollment_COBRAH_0.8.png}}\\[0.2cm]
\caption{Cumulative expected regret for several algorithms and budgets. \label{fig:case-study-results-regret}}
\end{figure}

 % We first compare the COBRAH, CUCB, CUCB-SW, and Random algorithms in terms of regret and enrollment across various budget (capacity) levels. 
 The plots in Figure~\ref{fig:case-study-results-regret}  so a regret comparison between the COBRAH, CUCB, CUCB-SW, and Random algorithms in using our case study data. These plots demonstrate that the COBRAH algorithm consistently outperforms other methods at all capacity levels. Specifically, COBRAH achieves sublinear regret across all budgets, which implies that the selected participants overlap significantly with the participants in the optimal super-arm at each period. Conversely, the other three algorithms repeatedly pick a high fraction of non-optimal participants, resulting in linear regret. This suggests that the comparison algorithms struggled to learn the expected rewards for each arm even after several rounds. 
%In terms of the average expected reward at each round, COBRAH also displays a dominant performance compared to the comparison algorithms. There is an interesting pattern what the expected average reward of COBRAH first increases from 16 to a maximum around 57 and then decreases to around 48. This is because initially when the budget is small, patients with highest enrollment probability which is close to 1 are picked, and the average expected reward is about the same as allowed number of patients. Under this scenario, average expected reward per round increases proportionally to the budget. As the budget limit increases, patients with lower probability of enrolling are enrolled, so the average expected reward decreases. 
With regard to patient enrollment, COBRAH maintains a significantly higher enrollment rate than the comparison algorithms. For instance, under a budget of 5\%, it achieves an average enrollment level (excluding the initial 30 periods) of 82.94\%, compared to 20.10\% for the Random policy, 14.48\% for CUCB, and 3.36\% for CUCB-SW. This results in a 312.64\% greater enrollment with respect to the next best algorithm (Random), highlighting that, even with stringent budgets, COBRAH can identify behavioral patterns that allow it to prioritize participants effectively, a determining factor in distinguishing successful interventions from less effective ones.
%Considering the effect of the budget limit on different algorithms,  COBRAH demonstrates pronounced effectiveness under stringent budget constraints (20\% and 40\%), where intelligent prioritization is most crucial due to limited available resources. In these scenarios, the performance gap between COBRAH and other methods is significantly amplified, indicating that its decision-making efficiency becomes especially beneficial when resources are scarce. 
Looking at the change in enrollment as the budget increases, the conversion rates for the comparison algorithms are fairly low. 
%As the available budget increases (60\% and 80\%), while all algorithms naturally improve due to additional resources, COBRAH maintains a clear advantage, emphasizing its robustness and superior adaptive decision-making. 
The relative improvements seen in CUCB and CUCB-SW under higher budgets (i.e., 40\% and 80\%) remain insufficient to match COBRAH’s performance, highlighting ongoing limitations in their ability to manage the complexities of agents with changing reward distributions effectively.  %\textbf{add description on random selection}
% Do we need to explain that data was from urban slums, but we are applying it to the office setting?

% We begin by going over the cumulative regret and average long-run net rewards for each of the four algorithms for a capacity level of 20\%. Figure~\ref{fig:cum-regret-ff-20} shows clearly that our algorithm has significantly lower cumulative regret than the random policy, CUCB, and SW-UCB algorithms. Notably, the random policy outperforms both CUCB and SW-UCB, suggesting that these algorithms are ill-fitted to the nonstationary setting. In practice, this figure shows that the KL-CUCB-FF policy makes smarter appointment allocations which lead to maximizing the central decision maker's rewards. 
\begin{figure}[ht]
\centering
\subfloat[Enrollment at 5\% capacity \label{fig:case-study-enrollment-5}]{\includegraphics[width=0.32\textwidth, trim={0 0cm 1.2cm 1.4cm},clip]{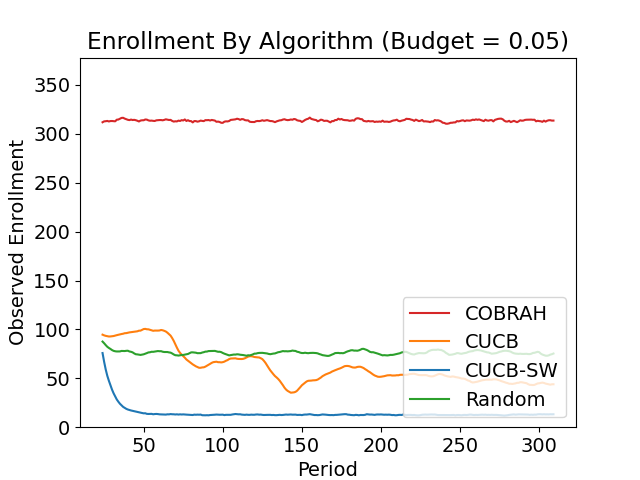}}
\subfloat[Enrollment at 10\% capacity \label{fig:case-study-enrollment-10}]{\includegraphics[width=0.32\textwidth, trim={0 0cm 1.2cm 1.4cm},clip]{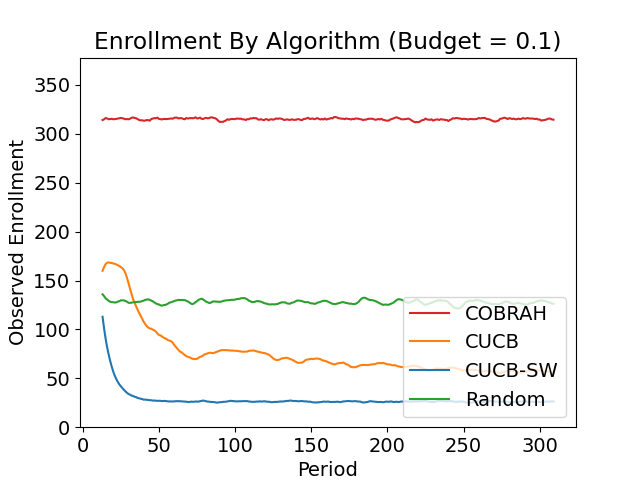}}
\subfloat[Enrollment at 20\% capacity \label{fig:case-study-enrollment-20}]{\includegraphics[width=0.32\textwidth, trim={0 0cm 1.2cm 1.4cm},clip]{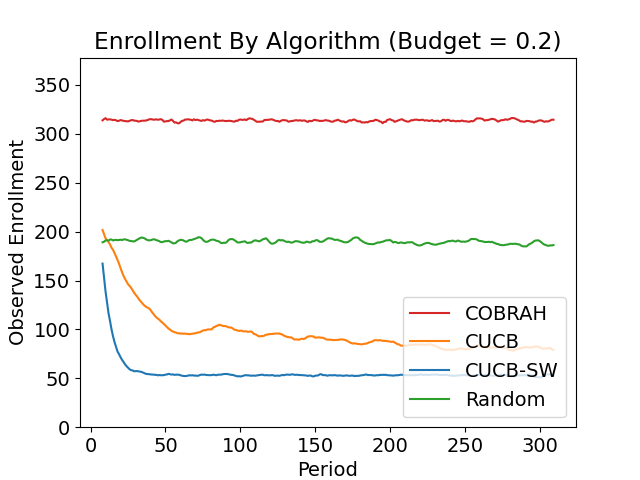}}\\[-1ex]
\subfloat[Enrollment at 40\% capacity \label{fig:case-study-enrollment-40}]{\includegraphics[width=0.32\textwidth, trim={0 0cm 1.2cm 1.4cm},clip]{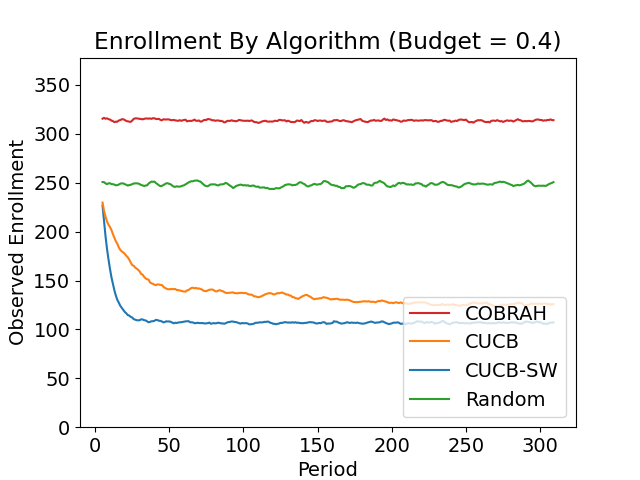}}
\subfloat[Enrollment at 80\% capacity \label{fig:case-study-enrollment-80}]{\includegraphics[width=0.32\textwidth, trim={0 0cm 1.2cm 1.4cm},clip]{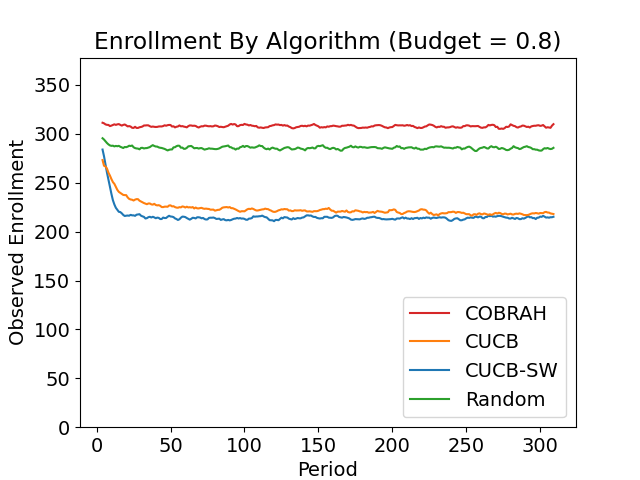}}\\[0.2cm]
\caption{Observed enrollment with five-period rolling average for several algorithms and budgets. \label{fig:case-study-results-enrollment}}
\end{figure}
\begin{figure}[ht]
\label{plot:vist_count}
\centering
\subfloat[CUCB\label{fig:hist-visit-count-cucb}]{\includegraphics[width=0.49\textwidth]{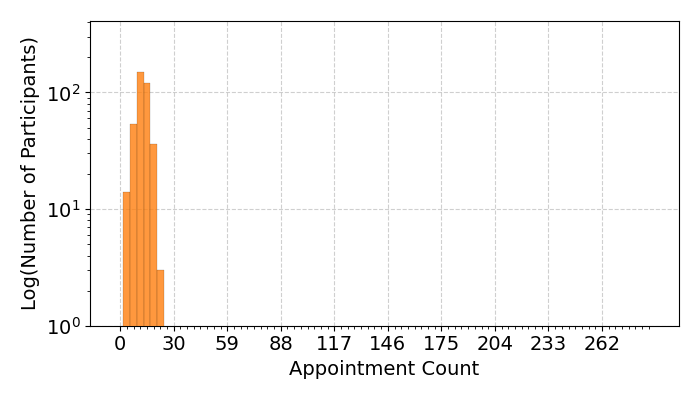}}
\subfloat[CUCB-SW\label{fig:hist-visit-count-cucb-sw}]{\includegraphics[width=0.49\textwidth]{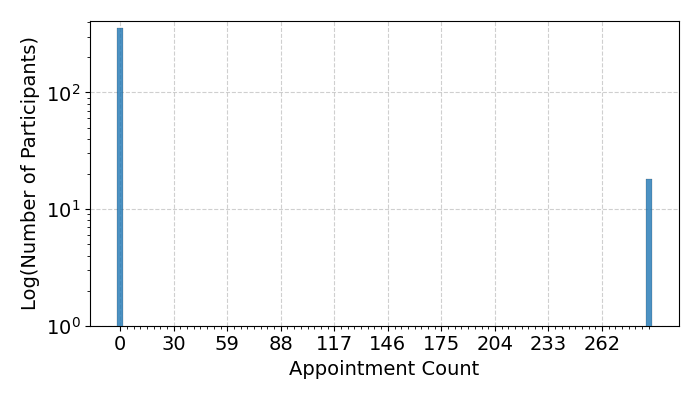}}\\[-1ex]
\subfloat[Random\label{fig:hist-visit-count-random}]{\includegraphics[width=0.49\textwidth]{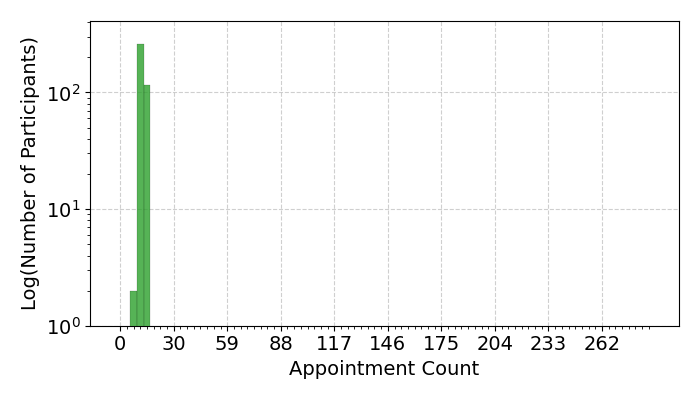}}
\subfloat[COBRAH\label{fig:hist-visit-count-cobrah}]{\includegraphics[width=0.49\textwidth]{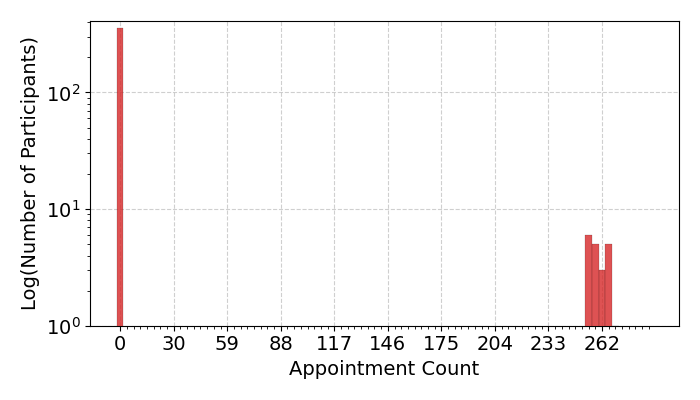}}\\[0.2cm]
\caption{Histograms of appointment count distributions for different algorithms and 20\% budget. Initialization rounds are omitted to highlight differences in allocation decisions for each algorithm. \label{fig:hist-visit-count}}
\end{figure}

Next, we examine differences in appointment allocation patterns to better understand the factors contributing to the varying performance across different algorithms.
%After this analysis on the three performance metrics across different budget limits, we delve deeper into individual patient-level visit policies, focusing on the 20\% budget scenario where COBRAH demonstrates its greatest advantage. 
Figure~\ref{fig:hist-visit-count} is an illustrative example with a budget of 5\% showing the logarithm of the number of participants that received each number of appointments (not including the initialization rounds). This figure allows us to examine how each algorithm distributes appointments among the participants across the planning horizon. CUCB and Random display a similar pattern, with visits more evenly distributed between all employees. On the other hand, CUCB-SW and COBRAH choose not to allocate any appointments to a large proportion of participants in favor of focusing on a subset of them. CUCB-SW chooses to visit this small subset of participants as much as possible, while COBRAH shows variation in the number of appointments offered to the participants chosen to be seen. While this distinction may appear marginal, the inferior regret and enrollment performance of CUCB-SW demonstrates that such subtle differences have meaningful implications. In the case of CUCB-SW, there is no further personalization, with the same subset of participants having appointments allocated at every period, while COBRAH has a more nuanced pattern that underscores its ability to adapt to changing motivational states. 
%The distribution of random choice is bell-shaped normal distribution which signifies the correct implementation of the method. 
%Notably, SW-UCB repeatedly visits the same subset of patients, leading to its highest regret and lowest enrollment. UCB1 has a long tail distribution which means that it pays quite frequent visits to a few patients. While it seems to distinguish the group of patients that need more patients, it cannot identify a more tailored group classification inside those patients. 
%In contrast, COBRAH strategically allocates visits, effectively categorizing patients into four distinct groups based on their visitation needs, ranging from infrequent to frequent visits. This nuanced classification underscores COBRAH’s ability to dynamically adapt to changing patient states and motivation levels.

\begin{figure}
    \centering
    \subfloat[Random\label{fig:visit-interval-random-0.05}]{\includegraphics[width=0.32\textwidth, trim={0cm 0cm 0cm 1.3cm},clip]{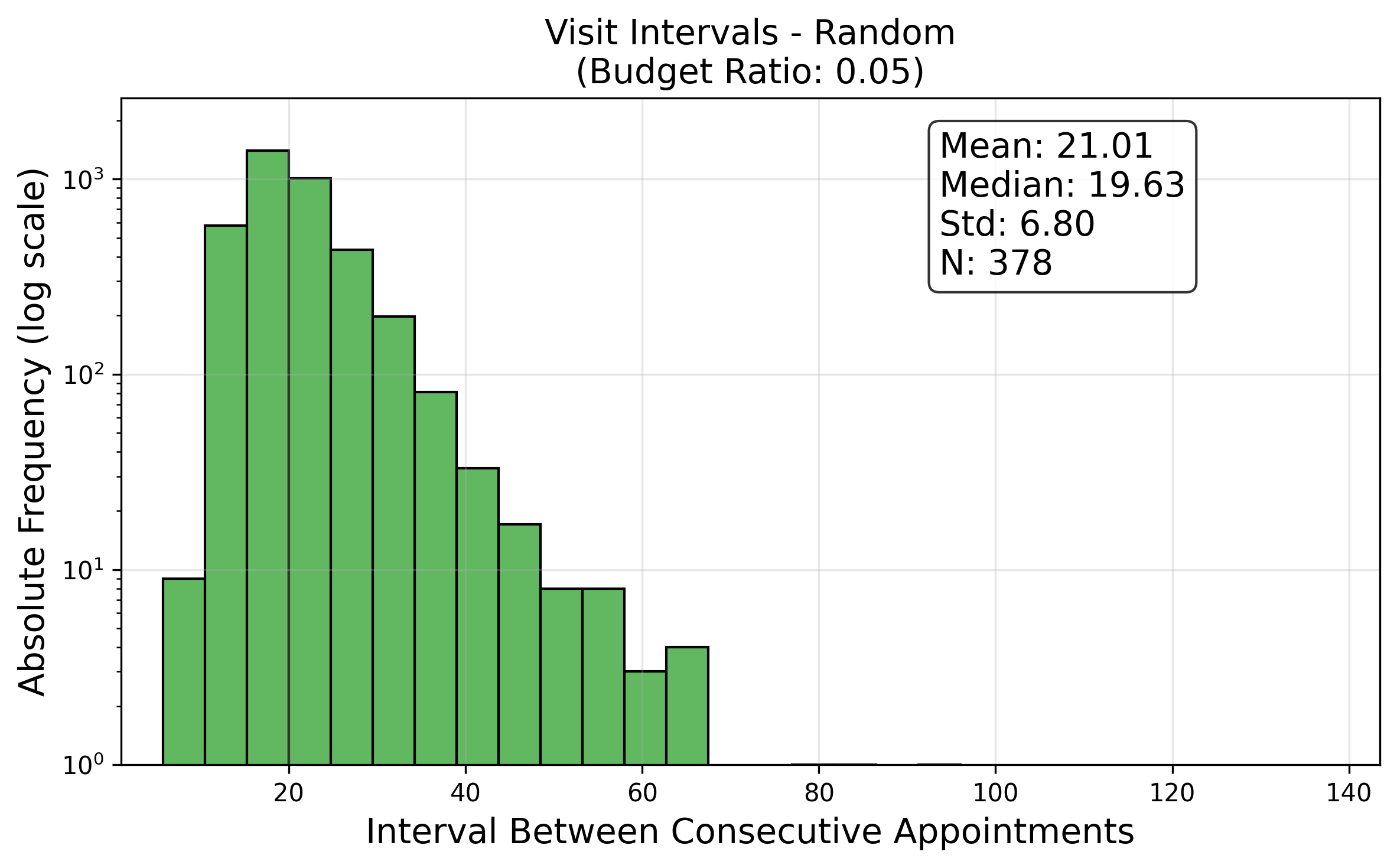}}
    \subfloat[CUCB\label{fig:visit-interval-cucb-0.05}]{\includegraphics[width=0.32\textwidth, trim={0cm 0cm 0cm 1.3cm},clip]{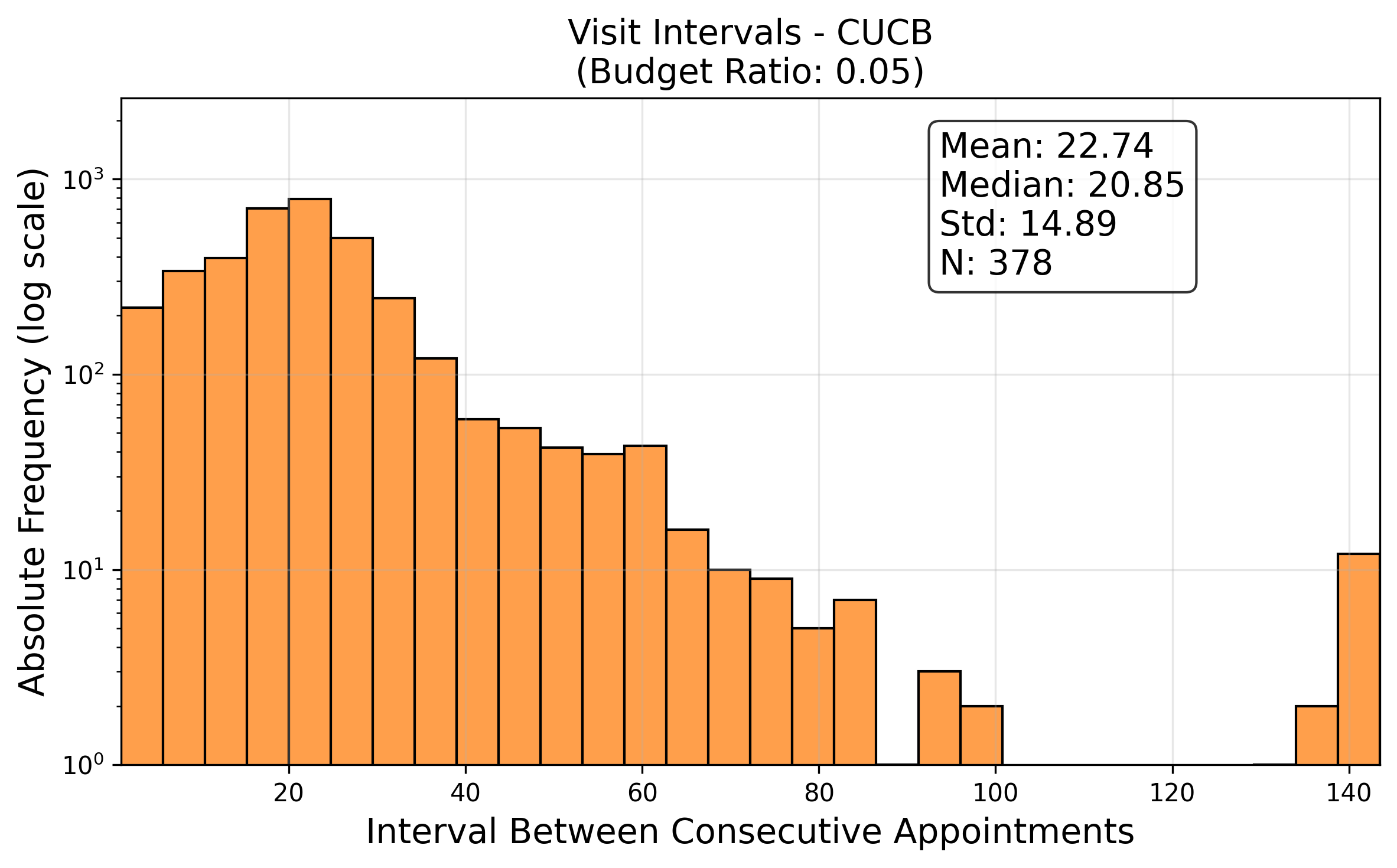}}
    \subfloat[COBRAH\label{fig:visit-interval-cobrah-0.05}]{\includegraphics[width=0.32\textwidth, trim={0cm 0cm 0cm 1.3cm},clip]{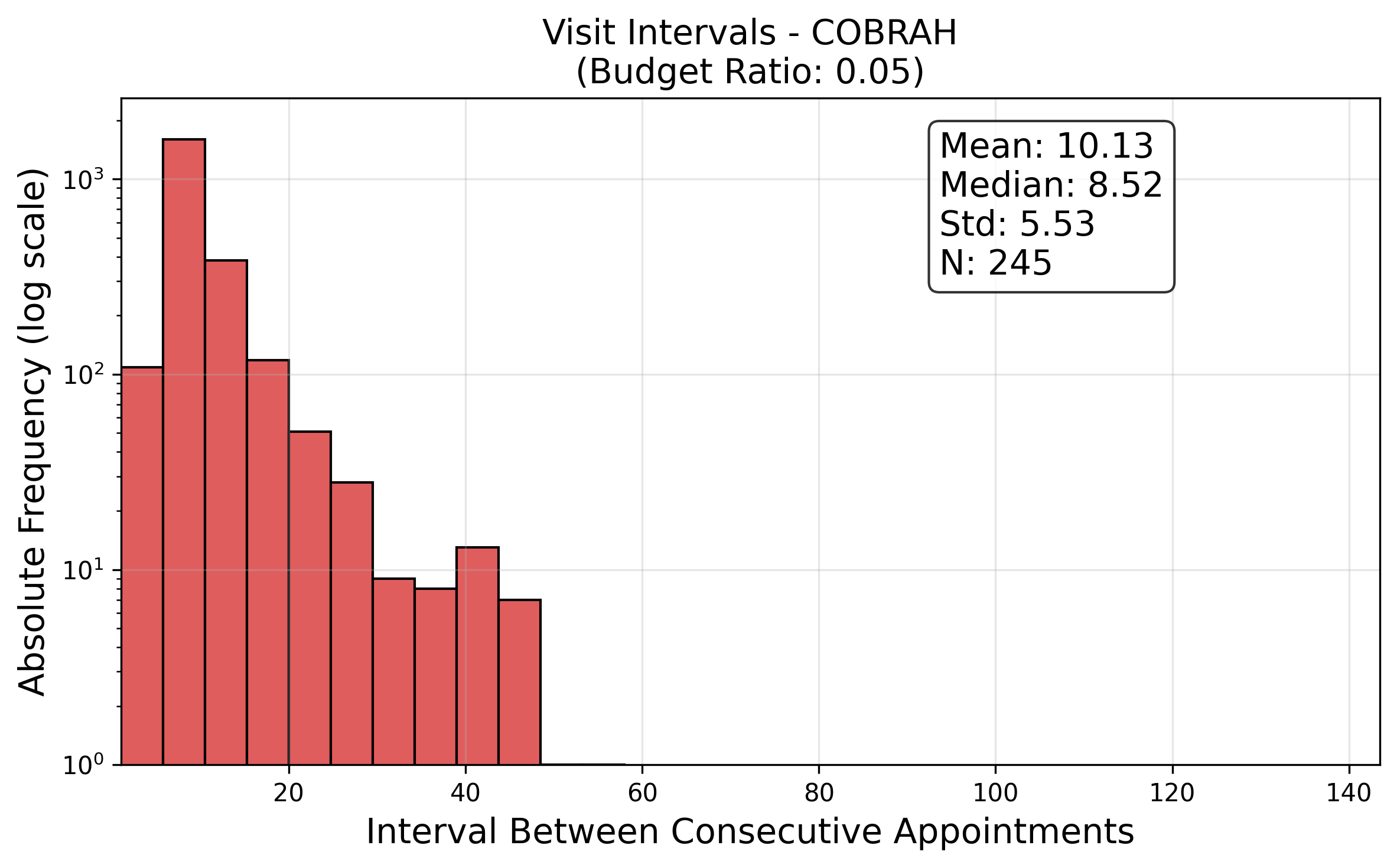}}\\[0.2cm]
    \caption{Histograms of visit intervals between consecutive appointments for different algorithms under 5\% budget. Participants with at least two visits after initialization rounds are included. \label{fig:visit-interval}}
\end{figure}

Figure~\ref{fig:visit-interval} offers additional insight into appointment allocation strategies by examining the histograms and summary statistics of intervals between consecutive appointments for the Random, CUCB, and COBRAH algorithms. The distribution is omitted for CUCB-SW because the same group of participants is seen at every period, leading to appointment intervals equal to one for all 18 participants seen. Compared to the Random and CUCB policies, COBRAH focuses on a subset of participants (245 out of 378) after the initialization rounds due to the low budget (5\%), leading to more frequent appointments both in terms of the mean and median. Furthermore, for the participants seen, the intervals between consecutive appointments are tailored based on individual needs, ranging from 2 to 57 periods.
%The trend from SW-UCB to random choice reveals a progressively broader and less focused distribution of visits, further highlighting the targeted and efficient nature of COBRAH’s approach. 

\section{Managerial Implications and Extensions}
\label{sec:exp-results-managerial-imp}

% \ka{Outline:
% \begin{itemize}
%     \item Value of NS bandit framework for behavioral health interventions
%     \item Discuss how FBG or other side information could be incorporated
%     \item Comment on this sentence that appears earlier in the paper: ``Finally, we note that our appointment allocation framework addresses a key challenge of successful PPP and CHW intervention implementation, which lies in programming, accountability, and performance assessment \citep{fanelli2020insights, kok2021community}."; Break down PPP argument explicitly
% \end{itemize}}

The experimental results in Sections~\ref{sec:exp-results} and \ref{sec:exp-results-case-study} show a compelling performance of the COBRAH algorithms for a class of problems where a central decision-maker seeks to drive agent behavior with no prior knowledge of the effect of their actions on each individual. We highlight three managerial implications of our proposed algorithms:
\begin{enumerate}
    \item \emph{Accounting for nonstationarity when reward distributions are subject to habituation and recovery dynamics leads to improved resource allocation decisions}. When central decision-makers seek to maximize a desired outcome by promoting behaviors in a target population, personalization and adaptation are crucial. As evidenced by our case study results, the COBRAH algorithms achieve up to 312.64\% greater average enrollment (excluding the first 30 periods) than the best comparison algorithm for the budgets considered. This improved efficiency is explained by the level of personalization and adaptation of the COBRAH algorithms, which appear both in terms of the number of appointments provided to each participant and the interval between consecutive visits. 
    \item \emph{Offering flexible solution approaches facilitates informed decision-making and enhances strategic competitiveness}.
    % alternatives: supports decision-makers in gaining a strategic upper hand; facilitates smarter decisions and strengthens strategic positioning
    The COBRAH algorithms provide finite-time guarantees for a general framework that is amenable to customization based on each decision maker's goals and context. Specifically, each central decision-maker selects their preferred reward measure, system dynamics (as long as standard modeling assumption are met), and may incorporate side information (if available) to more quickly improve the reward estimates for each arm. This side information can be integrated into our algorithms at the step where arm states and parameter estimates are computed. Beyond its ability to be tailored to each context, our framework is easy to adjust in future iterations as decision makers' strategic goals and priorities shift.
    \item \emph{The COBRAH algorithms provide a structured approach for CHW intervention planning}. In accordance with the WHO's recommendation on conducting research to optimize the design, implementation, and performance of CHW programs \citep{world2018guideline, kok2021community}, we developed a novel framework for CHW programming in the context of chronic disease care in resource-limited settings. Our framework allows central decision-makers to select CHW program staffing levels based on the desired community outcomes, build optimized plans for resource allocation decisions, and set realistic performance goals (e.g., via our regret bounds) that can be assessed periodically. These features of our framework are critical for settings that must rely on public-private-partnerships for program financing in that they clearly delineate the value of the financial investments made even when there is limited initial data available. 
\end{enumerate}

%%%%%% CONCLUSION %%%%%%
\section{Conclusion}

In this paper, we developed algorithms to efficiently plan the allocation of CHW appointments for diabetes care under partial information. Our framework can handle multiple tradeoffs, including between allocating appointments for care provision and information gathering. Our theoretical analyses show that our algorithms (COBRAH) achieve logarithmic regret in expectation. By developing a tuned version of our COBRAH-FF algorithm, we set the foundation for conducting simulation experiments and for practical implementation of our solution approaches.
%simulation experiments with real data from a partner company, we show that a slightly modified version of our algorithms significantly outperforms existing bandit algorithms \ka{(TO BE CONFIRMED)}. 
While our algorithms are not limited to healthcare settings, they are promising to improve personalized healthcare programs, especially for behavioral interventions that are typically affected by high attrition rates and high rates of undiagnosed conditions.

\bibliographystyle{informs2014.bst} % outcomment this and next line in Case 1tff
\bibliography{bibliography} 
\ECSwitch
%\ECDisclaimer
\ECHead{Electronic Companion}

\section{Proofs of Theoretical Results}\label{app:proofs}

\subsection{Proposition~\ref{prop:bad-round-increment}}
\proof{Proof of Proposition~\ref{prop:bad-round-increment}:}

Our result is an extension of existing work by \citep{mintz2020nonstationary}, who showed that the suboptimality gap of a bad round when a single arm is chosen is upper bounded by $L_g \textnormal{diam}(\Theta \times \mathcal{X})$, where $L_g$ is the Lipschitz continuity constant for the expected reward function $g(\theta,x)$ with respect to $\theta,x$ on compact set $\Theta \times \mathcal{X}$ and with diam$(\cdot)=\max_{x \in \mathcal{X}}\|x\|_2$. Due to the selection of up to $m$ arms per round in our problem, the suboptimality gap must be multiplied by $m$. We refer the reader to the proofs of Lemmas~1, 2, 3, and 4 in \cite{mintz2020nonstationary} for details.  \halmos
\endproof

\subsection{Proposition~\ref{prop-number-bad-rounds}}
\proof{Proof of Proposition~\ref{prop-number-bad-rounds}.}

The proof of this proposition consists of bounding the number of bad rounds separately for the case where our parameter estimates are close to their true values and the case where they are not. Let $T_{i,t}$ be the number of times arm $i$ has been pulled up to period $t$. We proceed by noting that, per Inequality~\eqref{eq:mintz-conc-ineq}, the average trajectory KL divergence for a single arm $i \in \{1,\dots,m\}$ exceeds $B(\alpha)\sqrt{\frac{\log(1/\alpha)}{T_{i,t-1}}}$ with probability no greater than $\alpha$.

We define the smallest optimality gap for each arm $i$ and across all arms as:
\begin{align}
    \Delta_{\textnormal{min}}^i &= \min_{t \in \mathcal{T}} \Bigg\{ 
\sum\limits_{j \in S_t^*} g(\theta_j, f_j^t(x_{j,0})) - \max_{S \in S_t^B: i \in S} \sum\limits_{j \in S} g(\theta_j, f_j^t(x_{j,0})) \Bigg\},\\
\Delta_{\textnormal{min}} &= \min_{i \in \{1,\dots,m\}} \Delta_{\textnormal{min}}^i
% \Delta_{\textnormal{max}}^i &= \min_{t \in \mathcal{T}} \Bigg\{ 
% \sum\limits_{j \in S_t^*} g(\theta_j, f_j^t(x_{j,0})) - \min_{S \in S_t^B: i \in S} \sum\limits_{j \in S} g(\theta_j, f_j^t(x_{j,0})) \Bigg\},
\end{align}
As a shorthand, let the true expectation of arm $i$ at time $t$ be $g_{i,t}=g\big(\theta_i, f_i^t(x_{i,0})\big)$ and its estimate be $\hat{g}_{i,t}=g\big(\hat{\theta}_i, f_i^t(\hat{x}_{i,0})\big)$, where the true expected reward ($g_{i,t}$) and the estimated expected reward ($\hat{g}_{i,t}$) are calculated after perpetuating the system dynamics using the true parameters/initial conditions of arm $i$ ($\theta_i$ and $x_{i,0}$) and its estimates ($\hat{\theta}_i$ and $\hat{x}_{i,0}$), respectively. We assume that the sequence of decisions applied to $(\theta_i, x_{i,0})$ and $(\hat{\theta}_i, \hat{x}_{i,0})$ is the same when applying the dynamics. Next, we provide some definitions that will be used throughout the proof.

We define the smallest average trajectory KL divergence between two arms that have an absolute difference in mean rewards of at least $\frac{\Delta_{\min}}{2m}$:
\begin{align}
    %\psi_i(\gamma) &= \max_{t \in \mathcal{T}} \Bigg\{\Big|g\big(\theta_i,f_i^t(x_{i,0})\big) - g\big(\hat{\theta}_i,f_i^t(\hat{x}_{i,0})\big)\Big|: \frac{1}{T_{i,t-1}} D_{i,\pi_i^t}(\theta_i,x_{i,0}||\hat{\theta}_i,\hat{x}_{i,0}) \leq \gamma \Bigg\}\\
    %\psi_{\max}(\gamma) &= \max_{i \in \mathcal{A}} \Bigg\{\max_{t \in \mathcal{T}} \bigg\{\Big|g\big(\theta_i,f_i^t(x_{i,0})\big) - g\big(\hat{\theta}_i,f_i^t(\hat{x}_{i,0})\big)\Big|: \frac{1}{T_{i,t-1}} D_{i,\pi_i^t}(\theta_i,x_{i,0}||\hat{\theta}_i,\hat{x}_{i,0}) \leq \gamma \bigg\}\Bigg\}\\
    % \delta_i &= \min_{t\in \mathcal{T}}\Bigg\{\min_{j \in A: j \neq i} \bigg\{\frac{1}{T_{i,t-1}} D_{i,\pi_1^t} (\theta_i,x_{i,0} || \theta_j,x_{j,0}) : |g(f_i^t(x_{i,0}),\theta_i) - g(f_j^t(x_{j,0}),\theta_j)| \geq \frac{\epsilon_i}{2} \bigg\}\Bigg\} \\
    % \delta_{\min} &= \min_{t \in \mathcal{T}}\Bigg\{\min_{i,j \in\mathcal{A}: i \neq j}\bigg\{\frac{1}{T_{i,t-1}} D_{i,\pi_1^t} (\theta_i,x_{i,0} || \theta_j,x_{j,0}) : |g(f_i^t(x_{i,0}),\theta_i) - g(f_j^t(x_{j,0}),\theta_j)| \geq \frac{\epsilon}{2} \bigg\}\Bigg\} \\
    \delta_{\min} &= \min_{t \in \mathcal{T}}\Bigg\{\min_{i,j \in \{1,\dots,m\}: i \neq j}\bigg\{\frac{1}{T_{i,t-1}} D_{i,\pi_1^t} (\theta_i,x_{i,0} || \theta_j,x_{j,0}) : |g_{i,t} - g_{j,t}| \geq \frac{\Delta_{\min}}{2m} \bigg\}\Bigg\}
\end{align}
We also define function $\psi_i(\gamma)$ to denote the maximum distance between the true and estimated rewards for arm $i$ when the average trajectory KL divergence is bounded by $\gamma$.
\begin{align}
    \psi_i(\gamma) &= \max_{t \in \mathcal{T}} \bigg\{\big|g_{i,t} - \hat{g}_{i,t}\big|: \frac{1}{T_{i,t-1}} D_{i,\pi_i^t}(\theta_i,x_{i,0}||\hat{\theta}_i,\hat{x}_{i,0}) \leq \gamma \bigg\}
\end{align}
Let $E_t = \Big\{\forall i \in \{1,\dots,m\}, \frac{1}{T_{i,t-1}}D_{i,\pi_1^t}(\theta_i,x_{i,0} || \hat{\theta}_i,\hat{x}_{i,0}) \leq \Lambda_{i,t}\Big\}$ and $\Lambda_{i,t}=B(t^{-4})\sqrt{\frac{4\log(t)}{T_{i,t-1}}}$. A bad round is defined as a round where we select a bad (suboptimal) super-arm $S_t \in \mathcal{S}_t^B$, where $\mathcal{S}_t^B \in \mathcal{S}\setminus \mathcal{S}_t^*$, where $\mathcal{S}_t^*$ denotes the set of optimal super-arms. Using these definitions, we can split the number of times we select a bad super-arm $T^B(n)$ as follows:
\begin{align}
    T^B(n) &= \sum_{t \in \mathcal{T}} \mathbbm{1}\{\Tilde{\pi}_t = S_t, S_t \in \mathcal{S}_t^B\} = \sum_{t \in \mathcal{T}} \mathbbm{1}\{\Tilde{\pi}_t = S_t, S_t \in \mathcal{S}_t^B, E_t\} + \sum_{t \in \mathcal{T}}\mathbbm{1}\{\Tilde{\pi}_t = S_t, S_t \in \mathcal{S}_t^B, \neg E_t\}
\end{align}
By Algorithm~\ref{alg:kl-cucb-sb} and the definition of $\psi_i(\gamma)$, we have that for all $i \in \{1,\dots,m\}, t \in \{1,\dots,n\}$:
\begin{align}
    \big| g^{\textnormal{UCB}}_{i,t} - \hat{g}_{i,t}\big| & \leq  \psi_i(\Lambda_{i,t})
\end{align}
For the case when $E_t$ holds, we also have that:
\begin{align}
    \big|g_{i,t} - \hat{g}_{i,t}\big| & \leq  \psi_i(\Lambda_{i,t})
\end{align}
Therefore we can establish a relationship between our upper confidence bounds and the true means:
\begin{align}
    |g_{i,t}^{\textnormal{UCB}} - g_{i,t}| &\leq 2 \psi_i(\Lambda_{i,t})
\end{align}
$E_t$ also implies that the upper confidence bounds hold:
\begin{align}
    g_{i,t}^{\textnormal{UCB}} - g_{i,t} &\leq 2 \psi_i(\Lambda_{i,t}) \nonumber\\
    g_{i,t}^{\textnormal{UCB}} &\leq  g_{i,t} + 2 \psi_i(\Lambda_{i,t}) \label{eq:true-ucb}
\end{align}
Based on the selection of super-arm $S_t\in \mathcal{S}_t^B$, there is at least one arm $i \in S_t$ for which $g^{\textnormal{UCB}}_{i,t} \geq g_{j,t}^{\textnormal{UCB}}$, where $j \in S_t^*, S_t^* \in \mathcal{S}_t^*$. Let $\Lambda_t = \max_{i \in S_t} \Lambda_{i,t}$ and $\psi_{\max}(\gamma)=\max_{i \in \{1,\dots,m\}} \psi_i(\gamma)$. Using the definition of $\psi_{\max}$ and summing over arms, we have:
\begin{align}
     \sum_{i \in S_t} g_{i,t} + 2m\cdot \psi_{\max}(\Lambda_{t}) & \myineq{\geq}{($i$)} \sum_{i \in S_t} \big(g_{i,t} + 2\cdot \psi_{\max}(\Lambda_{i,t})\big) \myineq{\geq}{($ii$)} \sum_{i \in S_t} \big(g_{i,t} + 2\cdot\psi_i(\Lambda_{i,t})\big) \nonumber \\
     &\myineq{\geq}{($iii$)} \sum_{i \in S_t} g_{i,t}^{\textnormal{UCB}} \myineq{\geq}{($iv$)} \sum_{j \in S_t^*} g_{j,t}^{\textnormal{UCB}} \myineq{\geq}{($v$)} \sum_{j \in S_t^*} g_{j,t} \label{eq:ineq-kl-cucb-sb}
\end{align}
where ($i$) follows from the definition of $\Lambda_t$, ($ii$) follows from the definition of $\psi_{\max}$, ($iii$) and ($v$) follow from the assumption that $E_t$ holds, and ($iv$) follows from the choice of super-arm $S_t$ by the COBRAH algorithm. Using the inequalities above, we have established a relationship between the rewards of the suboptimal and optimal super-arms:
\begin{align}
    \sum_{i \in S_t} g_{i,t} + 2m\cdot \psi_{\max}(\Lambda_{t}) &\geq \sum_{j \in S_t^*} g_{j,t}\\
    \psi_{\max}(\Lambda_{t}) &\geq \frac{1}{2m}\Bigg(\sum_{j \in S_t^*} g_{j,t} - \sum_{i \in S_t} g_{i,t} \Bigg)
\end{align}
By the definitions of $\Delta_{\min}$ and $\delta_{\min}$, we have that:
\begin{align}
    \psi_{\max}(\Lambda_{t}) &\geq \frac{\Delta_{\min}}{2m}\\
    \Lambda_{t} &\geq \delta_{\min}
\end{align}
Next we apply the definitions of $\Lambda_t$ and $\Lambda_{i,t}$: 
\begin{align}
    \max_{i \in S_t} \{\Lambda_{i,t}\} &\geq \delta_{\min}\\
    \max_{i \in S_t} \Bigg\{B(t^{-4})\sqrt{\frac{4\log(t)}{T_{i,t-1}}}\Bigg\} &\geq \delta_{\min}\\
    B(t^{-4})\sqrt{\frac{4\log(t)}{\min_{i \in S}T_{i,t-1}}} &\geq \delta_{\min}\\
    \min_{i \in S_t} T_{i,t-1} & \leq \frac{4\big(B(t^{-4})\big)^2\log(t)}{\delta_{\min}^2}
\end{align}
Therefore, when $E_t$ holds, the smallest number of times an arm $i \in S_t$ is pulled will not exceed the upper bound above. 
\begin{align}
    \sum_{t=1}^n \mathbbm{1}\{\Tilde{\pi}_t = S_t, S_t \in \mathcal{S}_t^B, E_t\} &= \sum_{t=1}^n \mathbbm{1}\Bigg\{\Tilde{\pi}_t = S_t, S_t \in \mathcal{S}_t^B, E_t, \min_{i \in S_t} T_{i,t-1} \leq \frac{4\big(B(m^{-4})\big)^2\log(t)}{\delta_{\min}^2}\Bigg\} \label{eq:klcucb-pi-bounds-hold}\\
    &\leq \sum_{t=1}^n \mathbbm{1}\Bigg\{\Tilde{\pi}_t = S_t, S_t \in \mathcal{S}_t^B, E_t, \min_{i \in S_t} T_{i,t-1} \leq \frac{4\big(B(m^{-4})\big)^2\log(n)}{\delta_{\min}^2}\Bigg\} \\
    &\leq \sum_{t=1}^n \sum_{i=1}^m \mathbbm{1}\Bigg\{\Tilde{\pi}_t = S_t, S_t \in \mathcal{S}_t^B, i 
\in S_t, E_t, \min_{i \in S_t} T_{i,t-1} \leq \frac{4\big(B(m^{-4})\big)^2\log(n)}{\delta_{\min}^2}\Bigg\}\\
    &\leq \frac{4m\big(B(m^{-4})\big)^2\log(n)}{\delta_{\min}^2}
\end{align}
where Equation~\eqref{eq:klcucb-pi-bounds-hold} follows from $t \geq m$ (at least $m$ rounds of initialization are required).

We now consider the case when $E_t$ does not hold.
\begin{align}
    \sum_{t=1}^n\mathbbm{1}\{\Tilde{\pi}_t = S_t, S_t \in \mathcal{S}_t^B, \neg E_t\} = \sum_{t=1}^n \mathbbm{1}\Bigg\{\Tilde{\pi}_t = S_t, S_t \in \mathcal{S}_t^B, \exists i \in S_t : T_{i,t-1} > \frac{4\big(B(t^{-4})\big)^2\log(t)}{\delta_{\min}^2}\Bigg\} \label{eq:ineq-bounds-hold-thm}
\end{align}
The event in the indicator function is a subset of the union of two events ($A$ and $B$):
\begin{align*}
    A &:= \Bigg\{\exists i \in S_t : g_{i,t}^{\textnormal{UCB}}- g_{i,t} > 2 \psi_i(\Lambda_{i,t})
    , T_{i,t-1} > \frac{4\big(B(t^{-4})\big)^2\log(t)}{\delta_{\min}^2}
    \Bigg\}\\
    B &:= \Bigg\{\exists j \in S_t^* : g_{j,t}^{\textnormal{UCB}} < g_{j,t}
    , T_{j,t-1} > \frac{4\big(B(t^{-4})\big)^2\log(t)}{\delta_{\min}^2}
    \Bigg\}
    % C &:= \{\exists i \in S_t, j \in S_t^* : g_{j,t} - g_{i,t} \leq 2 \psi_i(\Lambda_{i,t}), T_{i,t-1} > \ell_t\}
\end{align*}
where $A$ is the case where we severely overestimate the mean reward of a bad arm $i$ in a suboptimal super-arm $S_t \in \mathcal{S}_t^B$ and $B$ is the case where we severely underestimate the mean reward of an arm $j$ in an optimal super-arm $S_t^* \in \mathcal{S}^*_t$.
%, and $C$ is the case where the mean rewards for two arms (one in a suboptimal super-arm and another in the optimal super-arm) are close to each other. 
\begin{align*}
    \Bigg\{\Tilde{\pi}_t = S_t, S_t \in \mathcal{S}_t^B, \exists i \in S_t : T_{i,t-1} > \frac{4\big(B(t^{-4})\big)^2\log(t)}{\delta_{\min}^2}\Bigg\} \subseteq & \bigg\{\exists i \in S_t, s<t: \frac{1}{s}D_{i,\Tilde{\pi}_1^s} (\hat{\theta}_i, \hat{x}_{i,0}||\theta_i,x_{i,0})>\Lambda_{i,s}\bigg\} \cup\\
    & \bigg\{\exists j \in S_t^*, s'<t: \frac{1}{s'}D_{j,\Tilde{\pi}_1^{s'}} (\hat{\theta}_j, \hat{x}_{j,0}||\theta_j,x_{j,0})>\Lambda_{j,s}\bigg\}\\
    \subseteq & \bigcup_{1\leq s<t} \bigcup_{i \in S_t} \bigg\{\frac{1}{s}D_{i,\Tilde{\pi}_1^s} (\hat{\theta}_i, \hat{x}_{i,0}||\theta_i,x_{i,0})>\Lambda_{i,s}\bigg\} \cup\\
    & \bigcup_{1\leq s'<t} \bigcup_{j \in S_t^*} \bigg\{\frac{1}{s'}D_{j,\Tilde{\pi}_1^{s'}} (\hat{\theta}_j, \hat{x}_{j,0}||\theta_j,x_{j,0})>\Lambda_{j,s}\bigg\}
\end{align*}
Taking the expectation of the number of bad rounds per super-arm after $n$ rounds of play:
\begin{align}
    \mathbb{E}[T^B(n)] &\leq \frac{4\big(B(m^{-4})\big)^2\log(n)}{\delta_{\min}^2} + \mathbb{E}\Bigg[\sum_{t=1}^n \mathbbm{1}\bigg\{\Tilde{\pi}_t=S_t, S_t \in \mathcal{S}_t^B, \exists i \in S_t: T_{i,t-1}>\frac{4\big(B(t^{-4})\big)^2\log(t)}{\delta_{\min}^2}\bigg\}\Bigg] \\
    &\leq \frac{4\big(B(m^{-4})\big)^2\log(n)}{\delta_{\min}^2} + \sum_{t=1}^n \sum_{s=1}^{t-1} \sum_{s'=1}^{t-1} \sum_{i \in S_t} \sum_{j \in S_t^*} \Bigg[\mathbb{P}\bigg(\frac{1}{s}D_{i,\Tilde{\pi}_1^s} (\hat{\theta}_i, \hat{x}_{i,0}||\theta_i,x_{i,0})>B(s^{-4})\sqrt{\frac{4\log(s)}{T_{i,s-1}}}\bigg)+ \\
    &\hspace{211pt}\mathbb{P}\bigg(\frac{1}{s'}D_{j,\Tilde{\pi}_1^{s'}} (\hat{\theta}_j, \hat{x}_{j,0}||\theta_j,x_{j,0})>B((s')^{-4})\sqrt{\frac{4\log(s')}{T_{j,s'-1}}}\bigg)\Bigg] \label{eq:conc-ineq}\\
    &\leq \frac{4m\big(B(m^{-4})\big)^2\log(n)}{\delta_{\min}^2} + 2m^2 \sum_{t=1}^n \sum_{s=1}^{t-1} \sum_{s'=1}^{t-1} t^{-4} \\
    &\leq \frac{4m\big(B(m^{-4})\big)^2\log(n)}{\delta_{\min}^2} + 2m^2 \sum_{t=1}^n  t^{-2} \\
    &\leq \frac{4\big(B(m^{-4})\big)^2\log(n)}{\delta_{\min}^2} +  \frac{m^2\pi^2}{3} \label{eq:basel}
\end{align}
where Equation~\eqref{eq:conc-ineq} follows from Corollary~1 in \cite{mintz2020nonstationary} and Equation~\eqref{eq:basel} follows from the solution to the Basel Problem \citep{rockafellar2009variational}.
\halmos
\endproof

\subsection{Theorem~\ref{thm:klcucb-regret}}
\proof{Proof of Theorem~\ref{thm:klcucb-regret}:} The regret bound is obtained by calculating the maximum suboptimality gap at each time period multiplied by the expected number of bad rounds up to period $n$:
\begin{align}
    \E[R_{\Tilde{\pi}}(n)] &\leq \max_{t \in \mathcal{T}} \Bigg\{\sum\limits_{t=1}^n \sum\limits_{i \in S_t^*} g(\theta_i, f_{\pi^*}^t(x_{i,0})) -  \sum\limits_{t=1}^n \sum\limits_{i \in S} g(\theta_i, f_{\Tilde{\pi}}^t(x_{i,0}))\Bigg\} \cdot|\mathcal{S}|\cdot \E\big[ T^B(n)\big]\\
    &\leq \max_{t \in \mathcal{T}} \{\Delta_t\} \cdot |\mathcal{S}| \cdot \E\big[T^B(n)\big]
\end{align}
The final step of the proof follows from directly applying the results from Proposition~\ref{prop:bad-round-increment} and Proposition~\ref{prop-number-bad-rounds}. \halmos
\endproof

\subsection{Corollary~\ref{cor:regret-general}}
\proof{Proof of Corollary~\ref{cor:regret-general}.}

Following a similar process to the proof of Theorem~\ref{thm:klcucb-regret}, we begin by considering the case when $E_t$ holds. Note that because the total reward for pulling a super-arm is nonlinear, Equation~\eqref{eq:ineq-kl-cucb-sb} from the proof of Theorem~\ref{thm:klcucb-regret} will have to me modified. Let
\begin{align}
    \Bar{\psi}_i(\gamma) &= \max_{t \in \mathcal{T}}\bigg\{\max_{S \in \mathcal{S}: i \in S}\Big\{ \big|\bg_t(S) - \hat{\bg}_t(S)\big|: D_{i,\pi_i^t}(\theta_i,x_{i,0}||\hat{\theta}_i,\hat{x}_{i,0}) \leq \gamma \Big\}\bigg\}\\
    \Bar{\psi}_{\max}(\gamma) &= \max_{i \in \{1,\dots,m} \Bar{\psi}_i(\gamma)
\end{align}
For all $i\in \{1,\dots,m\}$ and for each period $t$ where $E_t$ holds:
\begin{align}
    r_{\bg_{t}}(S_t) + f\big(2\Bar{\psi}_{\max}(\Lambda_{t})\big) & \myineq{\geq}{($i$)} r_{\bg_{t}}(S_t) + f\big(2\Bar{\psi}_{\max}(\Lambda_{i,t})\big) \myineq{\geq}{($ii$)} r_{\bg_{t}}(S_t) + f\big(2\Bar{\psi}_{i}(\Lambda_{t})\big) \myineq{\geq}{($iii$)}  r_{\bg_{t}^\textnormal{UCB}}(S_t) \nonumber\\
    & \myineq{\geq}{($iv$)} r_{\bg_{t}^\textnormal{UCB}}(S_t^*) \myineq{\geq}{($v$)} r_{\bg_t}(S_t^*) \label{eq:ineqs-ff-bounds-hold}
\end{align}
where ($i$) follows from the definition of $\Lambda_t$, ($ii$) follows from the definition of $\Bar{\psi}_{\max}$, ($iii$) follows from Assumption~\ref{ass:bounded-smoothness}, ($iv$) follows from the selection of $S_t$ by Algorithm~\ref{alg:kl-cucb-ff}, and ($v$) follows from Assumption~\ref{ass:monotonicity} and $E_t$. We now have established a relationship between the expected rewards from choosing a suboptimal and optimal super-arm:
\begin{align}
    r_{\bg_{t}}(S_t) + f\big(2\Bar{\psi}_{\max}(\Lambda_{t})\big) & \geq r_{\bg_t}(S_t^*)\\
    f\big(2\Bar{\psi}_{\max}(\Lambda_{t})\big) & \geq r_{\bg_t}(S_t^*) - r_{\bg_{t}}(S_t)
\end{align}
% Let
% \begin{align}
%     \Bar{\Delta}_{\min} &= \min_{i \in 1,\dots,m}\bigg\{\min_{t \in \mathcal{T}} \Big\{r_{\bg_t}(S_t^*)-\max_{S_t \in \mathcal{S}_t^B: i\in S_t} r_{\bg_t}(S_t)\Big\}\bigg\}\\
%     % \Bar{\Delta}_{\min} &= \min_{i \in 1,\dots,m} \Delta_{\min}^i\\
%     \Bar{\delta}_{\min} &= \min_{t \in \mathcal{T}}\Bigg\{\min_{i,j \in \{1,\dots,m\}:i\neq j}\bigg\{\frac{1}{T_{i,t-1}}D_{i,\pi_1^t} (\theta_i,x_{i,0} || \theta_j,x_{j,0}) : |g_{i,t} - g_{j,t}| \geq \frac{f^{-1}\big(\Bar{\Delta}_{\min}\big)}{2}\bigg\}\Bigg\}
% \end{align}
% \ka{Perhaps $\bar{\delta}_{\min}$ and $\Bar{\psi}_i(\gamma)$ need to be redefined:
% \begin{align}
%     \bar{\delta}_{\min} &= \min_{t \in \mathcal{T}} \Bigg\{\min_{S, S' \in \mathcal{S}}\bigg\{\min_{i \in S, j \in S'}\Big\{D_{i,\pi_1^t} (\theta_i,x_{i,0} || \theta_j,x_{j,0}) : \big|\bg_{t}(S) - \bg_{t}(S')\big| \geq \frac{f^{-1}(\Bar{\Delta}_{\min})}{2}\Big\} \bigg\}\Bigg\}\\
%     \Bar{\psi}_i(\gamma) &= \max_{t \in \mathcal{T}}\bigg\{\max_{S \in \mathcal{S}: i \in S}\Big\{ \big|\bg_t(S) - \hat{\bg}_t(S)\big|: D_{i,\pi_i^t}(\theta_i,x_{i,0}||\hat{\theta}_i,\hat{x}_{i,0}) \leq \gamma \Big\}\bigg\}
% \end{align}}
Since $f(\cdot)$ is an invertible function:
\begin{align}
    \Bar{\psi}_{\max}(\Lambda_{t}) &\geq \frac{f^{-1}\big(r_{\bg_t}(S_t^*) - r_{\bg_{t}}(S_t)\big)}{2}\\
    \Bar{\psi}_{\max}(\Lambda_{t}) &\geq \frac{f^{-1}\big(\Bar{\Delta}_{\min}\big)}{2}\\
    \Lambda_t &\geq \Bar{\delta}_{\min} \label{eq:ff-tricky-proof-step}\\
    \max_{i \in S_t}\Bigg\{B(t^{-4})\sqrt{\frac{4\log(t)}{T_{i,t-1}}}\Bigg\} &\geq \Bar{\delta}_{\min}\\
    B(t^{-4})\sqrt{\frac{4\log(t)}{\min_{i \in S_t}T_{i,t-1}}} &\geq \Bar{\delta}_{\min}\\
    \min_{i \in S_t}T_{i,t-1} &\leq \frac{4\big(B(t^{-4})\big)^2 \log(t)}{\Bar{\delta}_{\min}^2}
\end{align}
where \eqref{eq:ff-tricky-proof-step} follows from the definition of $\Bar{\delta}_{\min}$.
The remaining steps of the proof are similar to the proof of Proposition~\ref{prop-number-bad-rounds} with events $A$ and $B$ are replaced by:
\begin{align}
    \Bar{A} & := \Bigg\{\exists i \in S_t : r_{\bg_t^{\textnormal{UCB}}}(S_t) - r_{\bg_t}(S_t) > f\big(2\bar{\psi}_i(\bar{\delta}_{i,t})\big), T_{i,t-1} > \frac{4\big(B(t^{-4})\big)^2 \log(t)}{\Bar{\delta}_{\min}^2} \Bigg\} \textnormal{ and}\\
    \Bar{B} & := \Bigg\{\exists j \in S_{t}^* : r_{\bg_t^{\textnormal{UCB}}}(S_t^*) < r_{\bg_t}(S_t^*), T_{j,t-1} > \frac{4\big(B(t^{-4})\big)^2 \log(t)}{\Bar{\delta}_{\min}^2} \Bigg\},
\end{align}
respectively.

\section{Case Study Details}

\subsection{NanoHealth Dataset}
\label{app:nanohealth-data}

\begin{figure}[ht]
    \centering
    \includegraphics[width=0.6\linewidth]{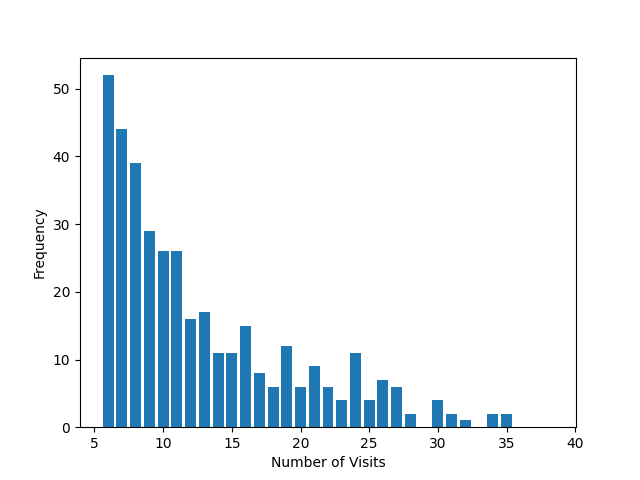}
    \caption{Distribution of number of visits per patient in the dataset provided by NanoHealth filtered by patients who had at least 6 visits.}
    \label{fig:nanohealth-data-visit-distribution}
\end{figure}

In this Appendix, we provide some details on the dataset that was used for our case study experiments described in Section~\ref{sec:case-study-exp-setup}. Specifically, Table~\ref{table:nanohealth-summary-data} provides summary statistics for the NanoHealth cohort, Table~\ref{table:patient_observations} provides summary statistics for visit information from NanoHealth's patient cohort, Table~\ref{table:nanohealth-summary-data-missing} presents the missingness for each feature, and Figure~\ref{fig:nanohealth-data-visit-distribution} shows the visit distribution for the dataset provided by NanoHealth.

\begin{table}[ht]
\begin{center}
\begin{tabular}{l c r}
\hline
Characteristic && \\ \hline
Number of patients && 378 \\ 
Age in years, mean (sd) && 53.3 (11.0) \\ 
Female, n (\%) && 219 (59.2) \\
Body mass index (kg/m$^2$), mean (sd) && 27.6 (5.9)\\
Waist circumference (inches), mean (sd) && 37.1 (4.3) \\
Heart rate (beats per minute), mean (sd) && 83.9 (12.3) \\ 
Diastolic blood Pressure (mmHg), mean (sd) && 90.2 (11.9) \\
Systolic blood Pressure (mmHg), mean (sd) && 138.2 (21.6) \\
Tobacco smoker, n (\%) && 29 (7.8)\\
Initial FBG (mg/dL), mean (sd) && 175.1 (71.9) \\
Final FBG (mg/dL), mean (sd) && 156.8 (61.6) \\
Number of management visits, mean (sd) && 13.1 (7.6) \\
Visits per month, mean (sd) && 0.6 (0.4) \\
\hline
\end{tabular}%}
\end{center}
\caption{Summary statistics from NanoHealth's patient cohort. See Table ~\ref{table:nanohealth-summary-data-missing} for the number of missing observations for each characteristic. \label{table:nanohealth-summary-data}}
\end{table}

\begin{table}[ht]
    \centering
    \begin{tabular}{l c}\hline
        Number of visits &  \\\hline
        Total across all patients & 5322\\
        Minimum per patient & 6 \\
        Maximum per patient & 39\\
        Average per patient & 14.1\\
        Standard deviation & 7.6\\\hline
    \end{tabular}
    \caption{Number of observations from NanoHealth's cohort.}
    \label{table:patient_observations}
\end{table}

\begin{table}[ht]
\begin{center}
\begin{tabular}{l c r}
\hline
Characteristic && Number of missing observations\\ \hline
Age in years&& 8 (2.1\%) \\ 
Female && 8 (2.1\%)\\
Body mass index && 133 (35.2\%) \\
Waist circumference  && 147 (38.9\%) \\
Heart rate  && 119 (31.5\%) \\ 
Diastolic blood Pressure  && 8 (2.1\%) \\
Systolic blood Pressure  && 18 (4.8\%) \\
Tobacco smoker && 8 (2.1\%)\\
Initial FBG  && 8 (2.1\%)\\
Final FBG  && 0 \\
Number of management visits && 0 \\
Visits per month && 0 \\
\hline
\end{tabular}%}
\end{center}
\caption{Missingness for NanoHealth's patient cohort. \label{table:nanohealth-summary-data-missing}}
\end{table}

% \subsection{Case Study Parameter Estimation}
% \label{app:param-est}

% \ka{THIS IS NOT THE CURRENT GRID SEARCH PROCEDURE, IT IS JUST MEANT TO FACILITATE WRITING THE CURRENT ONE. FEEL FREE TO EDIT/DELETE.}

% % GridSearch 
% \begin{algorithm2e}[h]
%     \DontPrintSemicolon
%     \KwData{$(\by, \bz, \bar{\bb}, S, B, G, R)$}
%     \KwResult{$w^*$, $x^*$ = ($\hat{\bb}^*$, $\bs^*$, $\btheta^*$, $p^*$, $\mu^*$, $\alpha^*$, $\theta_0^*$, $\lambda^*$, $s_0^*$, $\beta^*$, $\gamma^*$, $\rho^*$)}
%     \Begin{
%         $\Bar{w}= w^{\textnormal{UB}}$
        
%         $\Bar{x} = \emptyset$
        
%         searchGrid = list($S \times B \times G \times R$)
    
%         \For{$i \in |S \times B \times G \times R|$}{
%             $s_0^i$, $\beta^i$, $\gamma^i$, $\rho^i$ = searchGrid[$i$]
            
%             ($w$, $\hat{\bb}$, $\bs$, $\btheta$, $p$, $\mu$, $\alpha$, $\theta_0$, $\lambda$) = optProb($\by$, $\bz$, $\bar{\bb}$, $s_0^i$, $\beta^i$, $\gamma^i$, $\rho^i$)
            
%             \If{$w < \Bar{w}$}{
%                 $\Bar{w} = w$ 
                
%                 $\Bar{x}$ = ($\hat{\bb}$, $\bs$, $\btheta$, $p$, $\mu$, $\alpha$, $\theta_0$, $\lambda$, $s_0^i$, $\beta^i$, $\gamma^i$, $\rho^i$)
%             }      
%         }
        
%         $w^* = \Bar{w}$
        
%         $x^* = \Bar{x}$
%     }
%     \caption{Coarse grid search for parameter estimation. \label{alg:grid-search}}
% \end{algorithm2e}

\subsection{Additional Case Study Results}
\label{app:more-case-study-results}

% \begin{figure}[h!]
%     \centering
%     \begin{subfigure}[b]{0.45\textwidth}
%         \includegraphics[width=\textwidth]{figures/regretROGUE_0.1.png}
%         \caption{10\% capacity}
%         \label{fig:cum-regret-ff-10}
%     \end{subfigure}
%     \begin{subfigure}[b]{0.45\textwidth}
%         \includegraphics[width=\textwidth]{figures/regretROGUE_0.4.png}
%         \caption{40\% capacity}
%         \label{fig:cum-regret-ff-40}
%     \end{subfigure}
    
%     \begin{subfigure}[b]{0.45\textwidth}
%         \includegraphics[width=\textwidth]{figures/regretROGUE_0.6.png}
%         \caption{60\% capacity}
%         \label{fig:cum-regret-ff-60}
%     \end{subfigure}
%     \begin{subfigure}[b]{0.45\textwidth}
%         \includegraphics[width=\textwidth]{figures/regretROGUE_0.8.png}
%         \caption{80\% capacity}
%         \label{fig:cum-regret-ff-80}
%     \end{subfigure}
%     \caption{Cumulative regret for full feedback version of algorithms after 312 rounds of play with 378 arms and varying capacity levels.}
%     \label{fig:cum-regret-ff-others}
% \end{figure}

\begin{figure}[ht]
    \centering
    \subfloat[10\% capacity \label{fig:cum-regret-ff-10}]{\includegraphics[width=0.45\textwidth]{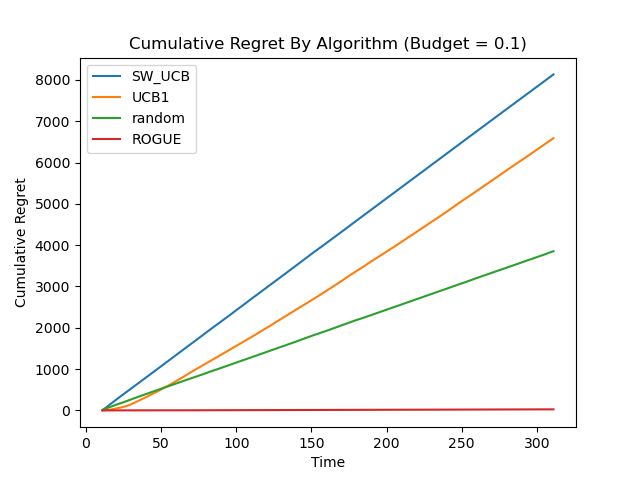}}
    \subfloat[40\% capacity \label{fig:cum-regret-ff-40}]{\includegraphics[width=0.45\textwidth]{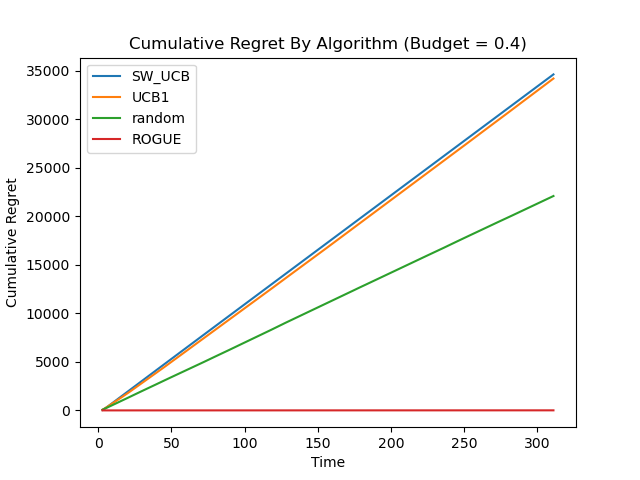}}
    
    \subfloat[60\% capacity \label{fig:cum-regret-ff-60}]{\includegraphics[width=0.45\textwidth]{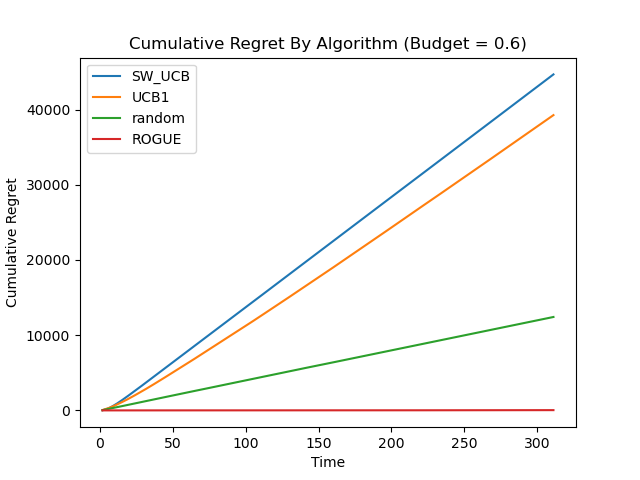}}
    \subfloat[80\% capacity \label{fig:cum-regret-ff-80}]{\includegraphics[width=0.45\textwidth]{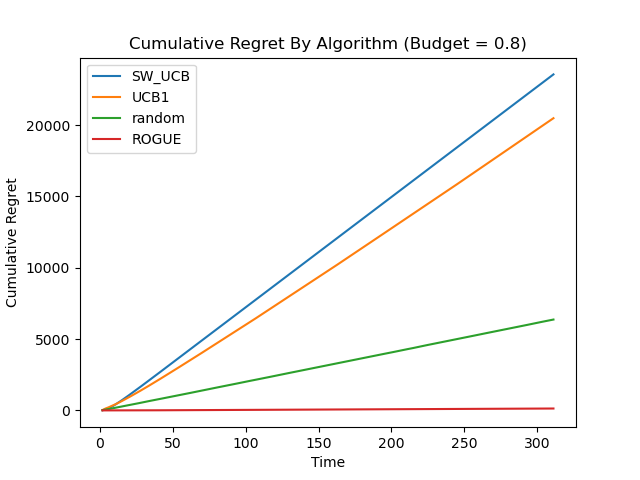}}
    \vspace{5pt}
\caption{Cumulative regret for full feedback version of algorithms after 312 rounds of play with 378 arms and varying capacity levels.\label{fig:cum-regret-ff-others}}
\end{figure}

\begin{figure}[ht]
    \centering
        \includegraphics[width=0.45\textwidth]{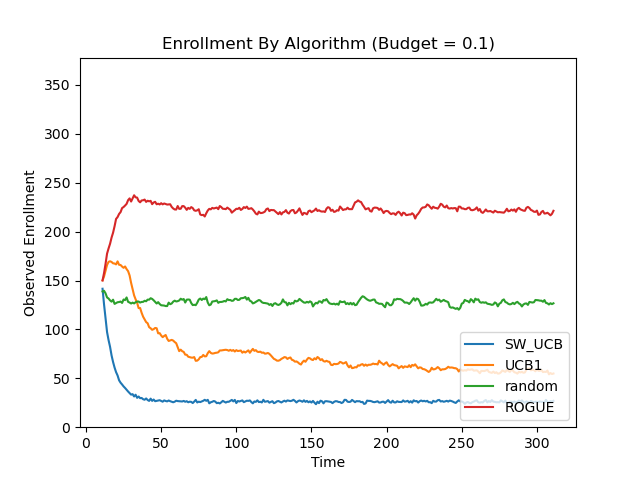}
        \includegraphics[width=0.45\textwidth]{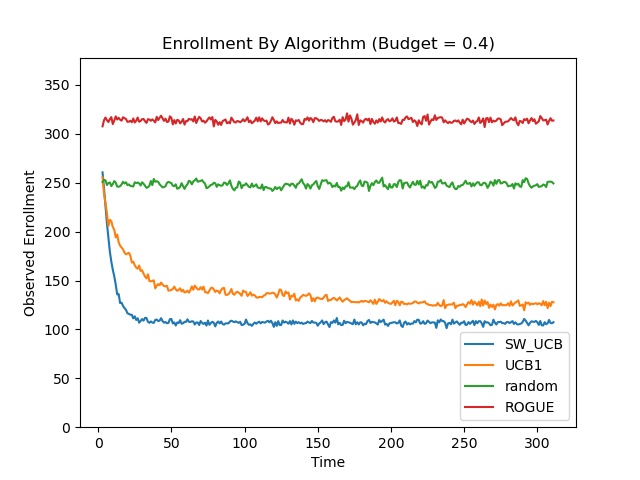}
        \includegraphics[width=0.45\textwidth]{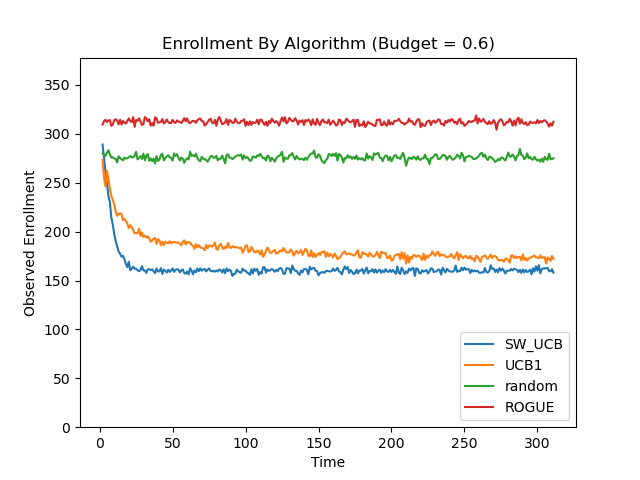}
        \includegraphics[width=0.45\textwidth]{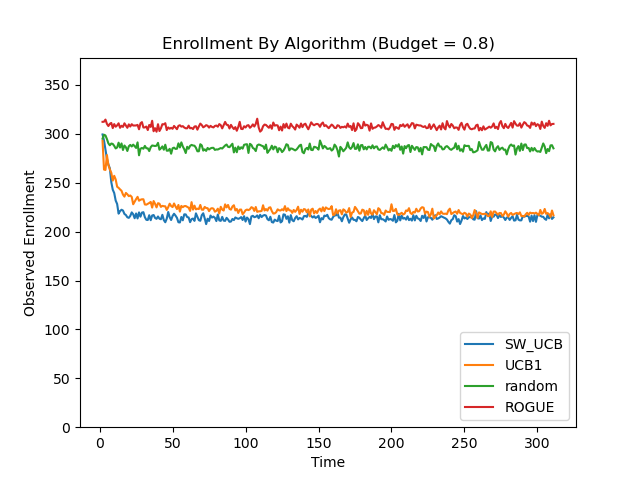}
\caption{Enrollment levels for full feedback version of algorithms after 312 rounds of play with 378 arms and capacity 10\% (upper left), 40\% (upper right), 60\% (lower left) and 80\% (lower right).\label{fig:enrollment-ff-others}}
\end{figure}

\end{document}